\documentclass[letterpaper]{article} 
\usepackage{aaai2026}  
\usepackage{times}  
\usepackage{helvet}  
\usepackage{courier}  
\usepackage[hyphens]{url}  
\usepackage{graphicx} 
\usepackage{natbib}  
\usepackage{caption} 
\usepackage{algorithmicx,algpseudocode,algorithm}

\usepackage{newfloat}
\usepackage{listings}
\DeclareCaptionStyle{ruled}{labelfont=normalfont,labelsep=colon,strut=off} 
\floatstyle{ruled}
\newfloat{listing}{tb}{lst}{}
\floatname{listing}{Listing}
\usepackage{amsmath,mathtools,amssymb}
\newcommand*{\dif}{\mathop{}\!\mathrm{d}}
\usepackage{booktabs}
\usepackage{multirow}
\usepackage{makecell}
\usepackage{subfig}

\title{Score-Based Model for Low-Rank Tensor Recovery}
\author{
    Zhengyun Cheng\textsuperscript{\rm 1},
    Changhao Wang\textsuperscript{\rm 1},
    Guanwen Zhang\textsuperscript{\rm 1}\thanks{Correspond author},
    Yi Xu\textsuperscript{\rm 2},
    Wei Zhou\textsuperscript{\rm 1},
    Xiangyang Ji\textsuperscript{\rm 3}
}
\affiliations{
    \textsuperscript{\rm 1}School of Electronics and Information, Northwestern Polytechnical University\\
   	\textsuperscript{\rm 2}School of Control Science and Engineering, Dalian University of Technology\\
   	\textsuperscript{\rm 3}Department of Automation, Tsinghua University\\
   	guanwen.zh@nwpu.edu.cn
}

\begin{document}
\maketitle
\begin{abstract}	
	Low-rank tensor decompositions (TDs) provide an effective framework for multiway data analysis. Traditional TD methods rely on predefined structural assumptions, such as CP or Tucker decompositions. From a probabilistic perspective, these can be viewed as using Dirac delta distributions to model the relationships between shared factors and the low-rank tensor.  However, such prior knowledge is rarely available in practical scenarios, particularly regarding the optimal rank structure and contraction rules. The optimization procedures based on fixed contraction rules are complex, and approximations made during these processes often lead to accuracy loss.
	To address this issue, we propose a score-based model that eliminates the need for predefined structural or distributional assumptions, enabling the learning of compatibility between tensors and shared factors. Specifically, a neural network is designed to learn the energy function, which is optimized via score matching to capture the gradient of the joint log-probability of tensor entries and shared factors.
	Our method allows for modeling structures and distributions beyond the Dirac delta assumption. Moreover, integrating the block coordinate descent (BCD) algorithm with the proposed smooth regularization enables the model to perform both tensor completion and denoising. Experimental results demonstrate significant performance improvements across various tensor types, including sparse and continuous-time tensors, as well as visual data.
\end{abstract}

\section{Introduction}
Tensor decompositions serve as powerful tools for analyzing high-order and high-dimensional data, aiming to capture the inter-dependencies among different modes by utilizing multiple shared factors. TDs have demonstrated remarkable success in various machine learning tasks, including data imputation \cite{zhe2016distributed, fang2021streaming}, factor analysis \cite{chen2022factor}, time-series forecasting \cite{miller2021tensor}, model compression \cite{novikov2015tensorizing, dai2023deep}, generative models \cite{glasser2019expressive, kuznetsov2019prior} among others. 

Mathematically, given an incomplete or noisy observation tensor $\hat{\mathcal{X}}$, the TDs model aims to recover unknown tensor $\mathcal{X}$ and noise component $\mathcal{S}$. Using Bayesian rule, the TDs model can be formulated as
\begin{equation}
	p(\mathcal{X},\mathcal{Z},\mathcal{S}|\hat{\mathcal{X}})\propto p(\hat{\mathcal{X}}|\mathcal{X},\mathcal{Z},\mathcal{S})
	p(\mathcal{S})
	p(\mathcal{X}|\mathcal{Z})
	p(\mathcal{Z}).
	\label{equ:bayes}
\end{equation}
The above equation holds because the noise $\mathcal{S}$ and the data $\mathcal{X}$ are independent.
Existing TD methods typically specify priors $p(\mathcal{Z})$ on latent factors $\mathcal{Z}$, and incorporate predefined contraction rules $\text{CR}(\mathcal{Z})$ and noise priors $p(\mathcal{S})$ to maximize the posterior probability of the observed tensor. This can be formulated as minimizing the negative log-posterior of Eq.~\eqref{equ:bayes}
\begin{equation}
	\begin{aligned}
		&-\log{p(\mathcal{X}, \mathcal{Z}, \mathcal{S}|\hat{\mathcal{X}})}\\
		=& -\log{p(\hat{\mathcal{X}}|\mathcal{X}, \mathcal{Z}, \mathcal{S})} &\Rightarrow& \|\hat{\mathcal{X}} - \mathcal{X} - \mathcal{S}\|_F &\text{(likelihood)} \\
		&- \log{p(\mathcal{S})} &\Rightarrow& \|\mathcal{S}\|_{\ell_1} &\text{(noise prior)} \\
		&- \log{p(\mathcal{X}|\mathcal{Z})} &\Rightarrow& \|\mathcal{X} - \text{CR}(\mathcal{Z})\|_F &\text{($\delta$-distri)} \\
		&- \log{p(\mathcal{Z})} &\Rightarrow& \mathfrak{R}(\mathcal{Z}) &\text{(factor prior)}
	\end{aligned}
\end{equation}
The second term models sparse or Laplacian noise, often regularized using the $\ell_1$-norm $\|\hat{\mathcal{X}} - \mathcal{X}\|_{\ell_1}$. The third term represents the contraction rule that defines the generative process from latent factors to the tensor, which is typically modeled as a Dirac delta distribution $p(\mathcal{X}|\mathcal{Z}) = \delta(\mathcal{X} - \text{CR}(\mathcal{Z}))$, and relaxed to an F-norm penalty $\|\mathcal{X}-\text{CR}(\mathcal{Z})\|_F$. The last term, $\mathfrak{R}(\mathcal{Z})$, serves as a low-rank regularizer.

From the perspective of modeling $p(\mathcal{X}|\mathcal{Z})$, the traditional approaches predominantly focus on identifying suitable contraction structures, including classical models such as CANDECOMP/PARAFAC (CP) \cite{kolda2009tensor}, Tucker \cite{zhou2015efficient}, and tensor train \cite{oseledets2011tensor}, along with their variants \cite{wang2017efficient, wu2022tensor}.
However, in real-world applications, such prior knowledge, e.g., the definition of tensor rank and contraction rules, is often unavailable. Moreover, effective low-rank models can vary significantly across different domains. Consequently, selecting an appropriate TD model for a specific dataset can be challenging.
Recent work, referred to as tensor network structure search \cite{li2020evolutionary, li2022permutation}, has demonstrated that choosing an appropriate contraction rule can substantially improve factorization performance. Another promising direction is learning nonlinear mappings directly from the observed tensor, using techniques such as nonparametric models \cite{chu2009probabilistic, xu2012infinite, zhe2016distributed} and deep neural networks \cite{liu2019costco, fang2021streaming, luo2023low, fan2021multi}. Empirical results show that nonlinear TD methods outperform traditional multilinear approaches in various applications, largely due to their greater expressive capacity.
Despite the success of nonlinear TDs in reducing structural assumptions, they still model the low-rank decompostion as the Dirac delta distribution conditioned on a parameterized contraction function. 

From the perspective of modeling the factor prior $p(\mathcal{Z})$, a Gaussian prior is often imposed on the latent factors, while the observed entries are modeled using a Gaussian likelihood \cite{rai2014scalable, zhao2015bayesian} or a Gaussian process \cite{xu2012infinite, zhe2016distributed}. These priors can also be realized through norm-based regularizations, such as the nuclear norm $\|\mathbf{X}\|_*$, the Schatten-$p$ norm $\|\mathcal{X}\|_{S_p}$ \cite{giampouras2020novel}, or the Frobenius norm of the factor tensors \cite{fan2023euclideannorminduced}.
However, in real-world applications, latent factors may originate from unknown or complex distributions, and observations can be generated through intricate, multi-modal processes. In the absence of knowledge about the true generative mechanism, such restrictive distributional assumptions can introduce model bias, limit the expressiveness of TD models, and ultimately lead to inaccurate estimations.

Recently, \cite{tao2023undirected} proposed to model the joint probability $p(\mathcal{X}, \mathcal{Z})$ of the tensor and its latent factors from an energy-based perspective. A variational Gaussian distribution was employed to approximate the true posterior of the factors, and a noise-contrastive loss was used for optimization. However, the method heavily relies on high-quality pairwise noise to avoid energy collapse and accurate posterior estimation, and it is not applicable to visual data.

\begin{figure}[t]
	\centering 
	\includegraphics[width=0.98\linewidth]{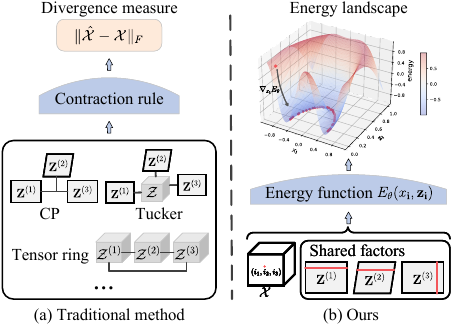}
	\caption{(a) The traditional method model the Dirac delta distribution of factors and tensors through predefined contraction rules. (b) From energy perspective, the proposed method model the gradient of the log joint probability density function with respect to the entry.} 
	\label{fig:overview}
\end{figure}
To address the above issues, this paper proposes to model the gradient of the joint probability distribution via a parameterized energy function, leveraging score matching with multiple noise levels for model optimization, as shown in Figure~\ref{fig:overview}. We capture low-rank structure through trainable shared factors on each mode, avoiding the limitations imposed by fixed contraction rules. The model is optimized using Adam, eliminating the need for complex iterative solvers such as Alternating Direction Method of Multipliers or Block Coordinate Descent commonly used in traditional tensor decomposition methods.
The proposed method offers several key advantages:
\begin{itemize}
	\item Our approach goes beyond the conventional modeling of $p(\mathcal{X}|\mathcal{Z})$ and $p(\mathcal{Z})$ separately, thereby relaxing the constraints imposed by the Dirac delta distribution. It features a flexible architecture that can adapt to various data distributions.
	\item By modeling the gradient of the joint probability, we avoid explicit posterior approximation of the factors and complex optimization procedures. Score matching enables direct and efficient model learning through gradient-based optimization.
	\item Integrating the model into an BCD framework allows for both tensor denoising and completion. Moreover, we proposed a smooth regularization from energy function perspective for denoising task.
	\item Extensive experiments on synthetic and real-world datasets demonstrate the significantly superior performance of the proposed method.
\end{itemize}

\section{Preliminaries}
\subsubsection{Notations.} We adopt similar notations with \cite{kolda2009tensor}. Throughout the paper, we use lowercase letters, bold lowercase letters, bold capital letters and calligraphic capital letters to represent scalars, vectors, matrices and tensors, e.g., $x,\mathbf{x},\mathbf{X}, \mathcal{X}$. Tensors refer to multi-way arrays which generalize matrices. For a D-order tensor $\mathcal{X}\in\mathbb{R}^{I_1\times\cdots\times I_D}$ , we denote its $(i_1, \cdots, i_D)$-th entry as $x_\mathbf{i}$.

\subsubsection{Denoising score matching with langevin dynamics.}
Let $p_\sigma(\tilde{x}|x):=\mathcal{N}(\tilde{x};x,\sigma^2)$ be a perturbation kernel, and $p_\sigma(\tilde{x}|x):=\int p_d(x)p_\sigma(\tilde{x}|x)\dif x$, where $p_d(x)$ denotes the data distribution. Consider a sequence of positive noise scales $\sigma_{min}=\sigma_1<\sigma_2<\cdots\sigma_L=\sigma_{max}$. Typically, $\sigma_{min}$ is small enough such that $p_{\sigma_{min}}(x)\approx p_d(x)$, and $\sigma_{max}$ is large enough such that $p_{\sigma_{max}}(x)\approx\mathcal{N}(x;0,\sigma^2_{max})$. \cite{song2019generative} propose to train a Noise Conditional Score Network, denoted by $\boldsymbol{s_\theta}(x,\sigma)$, with a weighted sum of denoising score matching \cite{vincent2011connection} objectives
\begin{multline}
	\boldsymbol{\theta}^*=\underset{\boldsymbol{\theta}}{\arg\min}\sum_{l=1}^L\left\{\sigma_l^2\mathbb{E}_{p_{\sigma_l}(\tilde{x}|x)p_d(x)}\right.\\
	\left.\left[\left\|\nabla_{\tilde{x}}\log p_\theta(\tilde{x},\sigma_l)-\nabla_{\tilde{x}}\log p_{\sigma_l}(\tilde{x}|x)\right\|_2^2\right]\right\}.
\end{multline}
Given sufficient data and model capacity, the optimal score-based model $\nabla_{\tilde{x}}\log p_{\theta^*}(\tilde{x},\sigma_l)$ matches $\nabla_{x}\log{p_\sigma(x)}$ almost everywhere for $\sigma\in\{\sigma_l\}^L_{l=1}$. For sampling, \cite{song2019generative} run $T$ steps of Langevin MCMC to get a sample for each $p_{\sigma_l}(x)$ sequentially
\begin{equation}
	x_l^t=x_l^{t-1}+\alpha_l\nabla_{\tilde{x}}\log p_{\theta^*}(x_l^{t-1},\sigma_l)+\sqrt{2\alpha_l}\epsilon_k,
\end{equation}
where $k=1,2,\cdots,K$, $\alpha_l>0$ is the step size, and $\epsilon_k$ is standard normal. The above is repeated for $l = L,L-1,\cdots,1$ in turn with $x^0_N\sim \mathcal{N}(x|0,\sigma_{max})$ and $x^0_l=x^K_{l+1}$ when $l<L$. As $K\rightarrow\infty$ and $\alpha_l\rightarrow 0$ for all $l$, $x^K_1$ becomes an exact sample from $p_{\sigma_{min}}(x)\approx p_d(x)$ under some regularity conditions.

\section{Methodology}
Given a noisy or incomplete observation tensor $\hat{\mathcal{X}} \in \mathbb{R}^{I_1 \times \cdots \times I_D}$, our method aims to recover the low-rank tensor $\mathcal{X}$, represented by $D$ shared factors $\mathcal{Z} = \{\mathbf{Z}^1, \dots, \mathbf{Z}^D\}$, and the noise tensor $\mathcal{S}$ when applicable for denoising tasks. We denote the factor associated with the $(i_1,\dots,i_D)$-th entry as $\mathbf{Z_i} = [\mathbf{z}^1_{i_1}, \dots, \mathbf{z}^D_{i_D}] \in \mathbb{R}^{D \times R}$, where $\mathbf{z}^d_{i_d} \in \mathbb{R}^{R}$ is the $i_d$-th row of $\mathbf{Z}^d$ and we use $\mathbf{z_i}$ to represent the flatten vectorization of $\mathbf{Z_i}$.
Instead of relying on the traditional intractable contraction rule and the factor priors to model $p(\mathcal{X}|\mathcal{Z})p(\mathcal{Z})$, we extend score-based methods to learn the first-order gradient of the log joint probability density function with respect to the tensor entry $x_\mathbf{i}$, conditioned on the shared factor according to its coordinate. Specifically, we model $\nabla_{x_\mathbf{i}} \log p(x_\mathbf{i}, \mathbf{z_i})$, referred to as the score at the tensor entry located in $\mathbf{i}=(i_1, \dots, i_D)$.
We further incorporate an BCD algorithm to enable the model to handle both tensor denoising and completion tasks effectively.

\subsection{Density Estimation with Denoising Score Matching for Tensor Entry}
It is worth noting that the score function from the perspective of tensor entries can be rewritten as:
$\nabla_{x_\mathbf{i}} \log p(x_\mathbf{i} \mid \mathbf{z_i}) 
= \nabla_{x_\mathbf{i}} \log \frac{p(x_\mathbf{i}, \mathbf{z_i})}{p(\mathbf{z_i})}
= \nabla_{x_\mathbf{i}} \log p(x_\mathbf{i}, \mathbf{z_i}).$
We model the joint distribution using an energy-based formulation:
\begin{equation}
	p(x_\mathbf{i}, \mathbf{z_i}) = \frac{e^{-E(x_\mathbf{i}, \mathbf{z_i})}}{\int e^{-E(x_\mathbf{i}, \mathbf{z_i})} \dif x_\mathbf{i} \dif\mathbf{z_i}},
\end{equation}
where $E(x_\mathbf{i}, \mathbf{z_i})$ denotes the joint energy function, and the denominator is the partition function that ensures the validity of the joint probability density function (JPDF). Following standard assumptions in tensor decomposition, we further assume independence among all tensor entries, with dependencies captured through shared latent factors.
To minimize the gradient while maximizing the curvature of the energy landscape around observation, we perturb tensor entries using multiple noise levels. For a given noise scale $\sigma$, the denoising score matching objective becomes:
\begin{multline}
	\label{equ:smObjective}
	\ell(\theta, \mathcal{Z}; \sigma) \triangleq \frac{1}{2} \mathbb{E}_{\tilde{x}_\mathbf{i} \sim \mathcal{N}(x_\mathbf{i}, \sigma^2),\, \mathbf{i} \sim \Omega} \\
	\left[ \left\| \frac{\tilde{x}_\mathbf{i} - x_\mathbf{i}}{\sigma^2} + \nabla_{\tilde{x}_\mathbf{i}} E_\theta(\tilde{x}_\mathbf{i}, \mathbf{z_i}) \right\|_F^2 \right],
\end{multline}
where the $-\nabla_{\tilde{x}_\mathbf{i}} E_\theta(\tilde{x}_\mathbf{i}, \mathbf{z_i}) = \nabla_{\tilde{x}_\mathbf{i}} \log p_\theta(\tilde{x}_\mathbf{i}, \mathbf{z_i})$, and $\Omega$ denotes the set of observed indices.
We then combine the objectives in Eq.~\eqref{equ:smObjective} across all noise scales $\sigma \in \{\sigma_l\}_{l=1}^L$ to form a unified loss:
\begin{equation}
	\mathcal{L}_D(\theta, \mathcal{Z}; \{\sigma_l\}_{l=1}^L) \triangleq \frac{1}{L} \sum_{l=1}^{L}\sigma_l^2 \, \ell(\theta, \mathcal{Z}; \sigma_l).
	\label{equ:lossfunction}
\end{equation}
Assuming the energy function $E_\theta(x_\mathbf{i}, \mathbf{z_i})$ has sufficient capacity, $E_{\theta^*}(x_\mathbf{i}, \mathbf{z_i})$ minimizes Eq.~\eqref{equ:lossfunction} if and only if $\|x_\mathbf{i}^* - x_\mathbf{i}\|_F < \sqrt{2 \eta \sigma_l}$ for all $l \in \{1, \dots, L\}$, where $x_\mathbf{i}^* = \arg\min_{x_\mathbf{i}} E_{\theta^*}(x_\mathbf{i}, \mathbf{z_i})$, so that $\nabla_{x_\mathbf{i}^*} E_{\theta^*}(x_\mathbf{i}^*, \mathbf{z_i}) = 0$, and $\eta$ represents the maximum residual error after optimization.

Therefore, instead of directly predicting a single most probable tensor from the observations, our method enables the model to capture the dependency between each tensor entry $x_\mathbf{i}$ and its corresponding shared factor $\mathbf{z_i}$, which act as conditions. Through score matching, the model automatically assigns lower energy values to correct answers and higher energies to corrupted or noisy ones. 

\subsubsection{Network architecture}
Effective modeling of probabilistic manifolds relies heavily on the network architecture. We define the energy function as $E_\theta(\tilde{x}_\mathbf{i}, \mathbf{z_i}) = g_{\theta_1}\big(g_2(g_{\theta_3}(\tilde{x}_\mathbf{i}), g_{\theta_4}(\mathbf{z_i}))\big)$,  where $g_{\theta_3}$ and $g_{\theta_4}$ are MLPs that encode $\tilde{x}_\mathbf{i}$ and $\mathbf{z_i}$, respectively, and $g_2$ is a summation or concatenation layer that couples tensor values with latent factors. The output layer is parameterized by $g_{\theta_1}$.
To model dynamic tensors, we incorporate a sinusoidal positional encoding layer \cite{tancik2020fourier}, denoted $\gamma_{\theta_t}(t_i)$, to capture temporal information. The energy function then becomes $E_\theta(\tilde{x}_\mathbf{i}, \mathbf{z_i}, t_i) = g_{\theta_1}\big(g_2(g_{\theta_3}(\tilde{x}_\mathbf{i}), g_{\theta_4}(\mathbf{z_i}), \gamma_{\theta_t}(t_i))\big)$. This architecture is widely used for temporal modeling and has been shown effective in capturing high-frequency patterns when combined with MLPs. In the above cases, the factor $\mathbf{z_i}$ is learnable parameter. While processing visual data, we employ a sinusoidal positional encoding layer to predict the shared factors, denoted $\mathbf{z_i} = \gamma_{\theta_\mathbf{i}}(\mathbf{i})$, and obtain an implicit representation.

\subsubsection{Posterior sampling}
\begin{algorithm}[t]
	\caption{Annealed Langevin dynamics for Tensor Recovery.}
	\label{algo:anneal}
	\textbf{Input}: Initialize $x_{\mathbf{i}}$.\\
	\textbf{Parameter}: $\{\sigma_l\}_{l=1}^L, \epsilon, K, \mathbf{i}\in\Omega$.\\
	\textbf{Output}: Optimized $\tilde{x}_{\mathbf{i}}$.
	\begin{algorithmic}[1]
		\For{$l \gets 1$ to $L$}
		\State{$\alpha_l\gets \epsilon \cdot \sigma_l^2/\sigma_L^2$} \Comment{$\alpha_l$ is the step size.}
		\For{$k \gets 1$ to $K$}
		\State{Draw $\epsilon_k \sim \mathcal{N}(0, 1)$}
		\State{$x_\mathbf{i}\gets x_{\mathbf{i}}-\alpha_l\nabla_{\tilde{x}_\mathbf{i}}E_{\theta^*}(x_\mathbf{i},\mathbf{z_i})+\sqrt{2\alpha_l}~\epsilon_k$}
		\EndFor
		\EndFor\\
		\Return{$x_{\mathbf{i}}$}
	\end{algorithmic}
\end{algorithm}
With the learned model parameters $\theta$ and factors $\mathcal{Z}$, the $\mathcal{X}$-subproblem can be formulated as 
\begin{equation}
	\label{equ:posteriorSampling}
	\arg\min_{x_\mathbf{i}}E_\theta(x_\mathbf{i}, \mathbf{z_i}), \forall\mathbf{i}\in\Omega.
\end{equation}
Unlike traditional TDs, direct predictions cannot be obtained even after learning the shared factors due to the utilization of an energy function. Instead, we need to seek for sampling methods of $p(x_\mathbf{i}|\mathbf{z_i})$. For Continuous data, we use score-based samplers by utilizing the score function, which is summarized as Algorithm~\ref{algo:anneal}. For handling discrete data like image data, we use gird search algorithm. 

\subsection{Smooth and Sparse Regularization for Denoising from Energy Perspective}
\begin{algorithm}[h]
	\caption{Block Coordinate Descent Algorithm for Tensor Denoising}
	\label{algo:tensor-BCD}
	\textbf{Input}: Observed tensor $\hat{\mathcal{X}} \in \mathbb{R}^{I_1 \times I_2 \times \cdots \times I_N}$ and index set $\Omega$, maximum iterations $T$, regularization parameter $\lambda_S$.\\
	\textbf{Output}: Estimated original tensor $\mathcal{X}^*$ and noise tensor $\mathcal{S}^*$.
	\begin{algorithmic}[1]
		\State Initialize $\mathcal{X}^{(0)} \gets \hat{\mathcal{X}}$
		\State Initialize $\mathcal{S}^{(0)} \gets \mathbf{0}$
		\For{$t = 1, 2, \ldots, T$}
		\For{$\mathbf{i} \in \Omega$}
		\State $x_\mathbf{i}^{(t)} \gets \arg\min_{x_\mathbf{i}} E_\theta(x_{\mathbf{i}}, \mathbf{z_i})$ 
		\Comment{Algorithm~\ref{algo:anneal}}
		\State $s_\mathbf{i}^{(t)}\gets\text{Soft}_{\lambda_S/2}(\hat{x}_\mathbf{i}-x_\mathbf{i}^{(t)})$
		\State $x_\mathbf{i}^{(t)}\gets\hat{x}_\mathbf{i}-s_\mathbf{i}^{(t)}$
		\State Update $\theta$ and $\mathbf{z_i}$ by minimizing $\mathcal{L}_D$ in Eq.~\eqref{equ:lossfunction} and $\mathcal{L}_S$ in Eq.~\eqref{equ:smoothReg} \Comment{Adam optimizer}
		\EndFor
		\EndFor
		\State \Return $\mathcal{X}^* \gets fold(\{x_\mathbf{i}^{(T)}\})$, $\mathcal{S}^* \gets fold(\{s_\mathbf{i}^{(T)}\})$
	\end{algorithmic}
\end{algorithm}

Given the current estimation of energy function $E_\theta(\cdot,\cdot)$ and the learned factor $\mathcal{Z}$, we aim to obtain the complete tensor, conditional on the observed values, i.e., $p(\mathcal{X},\mathcal{S}|\hat{\mathcal{X}},\mathcal{Z})$, such that the estimated complete data $\mathcal{X}$ and $\mathcal{S}$ can be updated by taking block coordinate descent algorithm.  Using Bayesian rule, the problem can be formulated as
\begin{equation}
	p(\mathcal{X},\mathcal{S}|\hat{\mathcal{X}},\mathcal{Z})\propto p(\hat{\mathcal{X}}|\mathcal{X},\mathcal{S})p(\mathcal{S})p(\mathcal{X}|\mathcal{Z}).
\end{equation}
The above equation holds because $\mathcal{Z}$ is independent of $\mathcal{S}$ and $\hat{\mathcal{X}}$.
With the probability relationship, i.e., $p(\hat{\mathcal{X}}|\mathcal{X},\mathcal{S})\propto\exp{(-\frac{\|\hat{\mathcal{X}}-\mathcal{X}-\mathcal{S}\|^2_F}{2\sigma^2})}$, and the sparse noise prior $p(\mathcal{S})\propto\exp{(-\lambda\|\mathcal{S}\|_{\ell_1})}$, and we solve the problem by minimizing the negative log-posterior. Thus the $\mathcal{S}$-subproblem can be formulated as 
\begin{equation}
	\label{equ:sparseReg}
	\arg\min_{\mathcal{S}}\|\hat{\mathcal{X}}-\mathcal{X}-\mathcal{S}\|_F^2+\lambda_S\|\mathcal{S}\|_{\ell_1}.
\end{equation}
which can be exactly solved by soft-thresholding operator applied on each element of the input, i.e., $\mathcal{S}=\mathop{Soft}_{\lambda_S/2}(\hat{\mathcal{X}}-\mathcal{X})$, where the $\mathop{Soft}_{\frac{\lambda_S}{2}}(\cdot)=\mathop{sgn}(\cdot)\max(|\cdot|-\frac{\lambda_S}{2}, 0)$. 

It is challenging to effectively remove complex mixed noise patterns, such as Gaussian noise, stripe noise, and dead lines, using only sparse regularization terms. Smoothness regularization has proven to be a powerful technique for denoising tasks, with total variation (TV) loss being a representative example. However, reconstructing the entire tensor in each iteration is computationally expensive and limits scalability.
To address this issue and promote smoothness in the shared factor space while improving generalization, we propose an energy-based regularization strategy. Specifically, we introduce a smoothing loss that penalizes large variations in the energy function with respect to small perturbations in the shared latent factors. The objective is defined as:
\begin{equation}
	\label{equ:smoothReg}
	\mathcal{L}_S(\theta; t) = \mathbb{E}_{\epsilon \sim \mathcal{N}(0, \sigma_S^2\mathbf{I}),\, \mathbf{i} \sim \Omega}\left[E_\theta\left(x_\mathbf{i}^{(t)}, \gamma_{\theta_\mathbf{i}}(\mathbf{i}+\epsilon)\right)\right],
\end{equation}
where $\epsilon$ denotes a random perturbation sampled from an isotropic Gaussian distribution, $x_\mathbf{i}^{(t)}$ is the posterior sample obtained at iteration $t$, as defined in Eq.~\eqref{equ:posteriorSampling}, and $\mathbf{z}_{\mathbf{i}+\epsilon}=\gamma_{\theta_\mathbf{i}}(\mathbf{i}+\epsilon)$ represents the latent factor at the perturbed location. This formulation encourages the model to assign lower energy values to nearby configurations of the shared factors, thereby promoting local smoothness and significantly reducing computational overhead while maintaining high denoising performance.
To better illustrate the proposed method, we summarize the steps of our method for tensor denoising in Algorithm \ref{algo:tensor-BCD}.

\section{Experiments}
This section presents the experimental outcomes of our method, contrasting them with the current state-of-the-art methods, as assessed on synthetic and real-world tensor.

\subsection{Simulation study}
\begin{figure}[t]
	\centering
	\subfloat[Beta]
	{\includegraphics[width=0.33\linewidth]{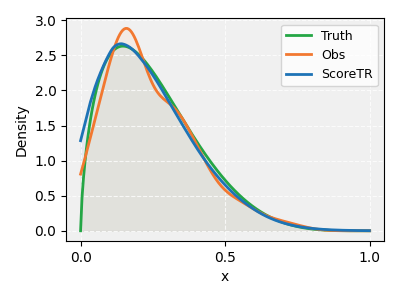}}
	\subfloat[MoG]
	{\includegraphics[width=0.33\linewidth]{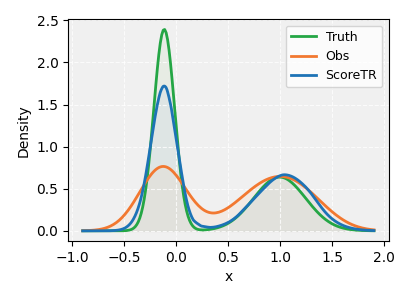}}
	\subfloat[Exponential]
	{\includegraphics[width=0.33\linewidth]{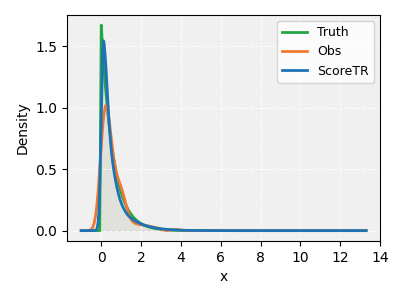}}
	\caption{Simulation results for different distributions.} 
	\label{fig:Simulation}
\end{figure}
\begin{figure}[t]
	\centering 
	\includegraphics[width=1.0\linewidth]{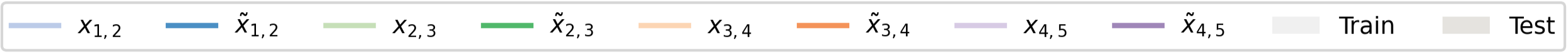}
	\vskip -0.3cm
	\subfloat[MR = 0.2]
	{\includegraphics[width=0.33\linewidth]{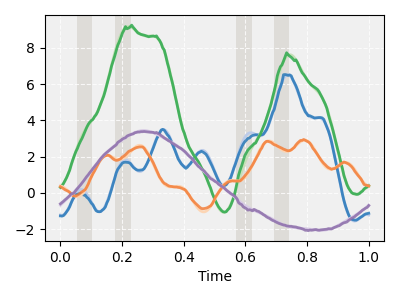}}
	\subfloat[MR = 0.4]
	{\includegraphics[width=0.33\linewidth]{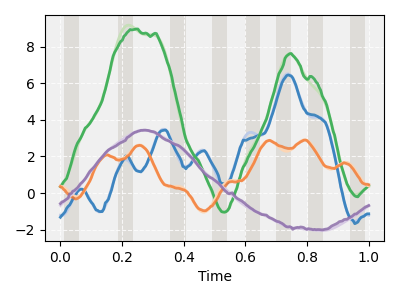}}
	\subfloat[MR = 0.6]
	{\includegraphics[width=0.33\linewidth]{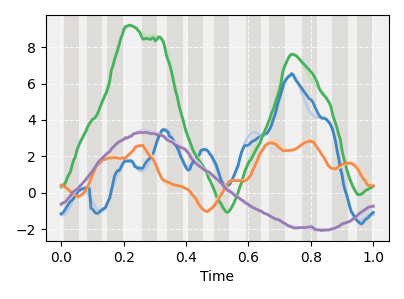}}
	\caption{Simulation results for continuous tensor recovery.} 
	\label{fig:SimulationC}
\end{figure}
\subsubsection{Tensors with non-Gaussian distributions}
Traditional TD methods often assume that tensor entries follow a Gaussian distribution. However, real-world data frequently exhibit more complex distributions. In this experiment, we evaluate the ability of our model to capture non-Gaussian distributions.
We consider a two-mode tensor of size $I \times I$, with $I = 8$. First, we generate two latent factor matrices of size $I \times R$, where the rank $R$ is set to 5. Then, conditioned on these factors, we generate tensor observations from three different non-Gaussian distributions: (1) Beta distribution, (2) Mixture of Gaussians (MoG), and (3) Exponential distribution. For each entry, we generate $N = 200$ independent samples.
For all settings, the latent factors are sampled independently and identically from a uniform distribution: $\mathbf{Z}^{1}, \mathbf{Z}^{2} \overset{\text{i.i.d.}}{\sim} \text{Uni}(0, 1)$.
For Beta distribution, each entry is sampled as  
$x_{ij} \overset{\text{i.i.d.}}{\sim} \text{Beta}\left((\mathbf{Z}^1 \mathbf{Z}^{2,\top})_{ij},\, 5\right).$
For Mixture of Gaussians (MoG), each entry follows  
$x_{ij} \overset{\text{i.i.d.}}{\sim} 0.6 \cdot \mathcal{N}\left(\cos((\mathbf{Z}^1 \mathbf{Z}^{2,\top})_{ij}),\, 0.1^2\right) + 0.4 \cdot \mathcal{N}\left(\sin((\mathbf{Z}^1 \mathbf{Z}^{2,\top})_{ij}),\, 0.25^2\right).$
For Exponential distribution, each entry is sampled as  
$x_{ij} \overset{\text{i.i.d.}}{\sim} \text{Exp}\left((\mathbf{Z}^1 \mathbf{Z}^{2,\top})_{ij}\right).$

We visualize the learned probability density function of a single tensor entry, as shown in Figure~\ref{fig:Simulation}. Our method does not simply fit the observed values, instead, it provides a more accurate approximation of the underlying true distribution. This demonstrates the flexibility of our model in capturing complex, non-Gaussian data distributions.

\subsubsection{Continuous tensor}
We next consider a dynamic tensor, where each entry corresponds to a time series. Following the setup in \cite{tao2023undirected}, we use a two-mode tensor of size $8 \times 8$, with each entry being a time series of length 200. We first generate latent factor matrices of size $8 \times 2$. Each row of the first factor matrix is sampled independently as $\mathbf{z}_\mathbf{i}^1 \sim \mathcal{N}([0, 2], 2 \cdot \mathbf{I})$, and each row of the second factor matrix is sampled as $\mathbf{z}_\mathbf{i}^2 \sim \mathcal{N}([1, 1], 2 \cdot \mathbf{I})$. Then, for each entry, we generate $N = 200$ observations over the time interval $t \in [0, 1]$. The tensor entries are computed using a weighted sum of temporal basis functions: $x_{\mathbf{i}}(t) = \sum_{r_1=1}^{2} \sum_{r_2=1}^{2} z_{i_1 r_1}^1 z_{i_2 r_2}^2 \omega_{r_1 r_2}(t)$, where $\omega_{11}(t) = \sin(2\pi t)$, $\omega_{12}(t) = \cos(2\pi t)$, $\omega_{21}(t) = \sin^2(2\pi t)$, and $\omega_{22}(t) = \cos(5\pi t)\sin^2(5\pi t)$. This synthetic dataset contains both low-frequency trends and high-frequency fluctuations, simulating realistic temporal dynamics. In addition to the fully observed case, we also evaluate performance under missing rates (MR) of 20\%, 40\%, and 60\%. Specifically, for each entry, we randomly select 4, 8, and 12 starting time points and mark the subsequent 5\% of the time stamps as missing.

Figure~\ref{fig:SimulationC} shows the completion results, plotting the learned trajectories of four tensor entries. It can be observed that higher missing rates lead to a slight performance decline. Our method is capable of adapting to complex scenarios that involve both low-frequency trends and high-frequency components, achieving near-perfect predictions.

\subsection{Tensor Completion}
We evaluate our model on tensor completion tasks using two sparse and two dynamic tensors.
\begin{table}[t]
	\centering
	\small
	\caption{Quantitative results of sparse tensor completion.}
	\label{tab:sparseTC}
	\setlength{\tabcolsep}{0.6mm}{
		\begin{tabular}{c|cccc|cccc}
			\toprule
			\multicolumn{1}{c}{Metric} &\multicolumn{4}{c}{RMSE} & \multicolumn{4}{c}{MAE} \\
			\midrule
			Method &R3 &R5 &R8 &R10 &R3 &R5 &R8 &R10 \\
			\midrule
			\multicolumn{9}{c}{\textit{Alog} $(200\times 100\times 200)$}\\
			\midrule
			CP-WOPT 
			&1.486 &1.386 &1.228 &1.355 &0.694 &0.664 &0.610 &0.658 \\
			GPTF 
			&0.911 &0.867 &0.878 &0.884 &0.511 &0.494 &0.530 &0.554 \\
			HGP-GPTF 
			&0.896 &0.867 &0.850 &0.844 &0.479 &0.473 &0.474 &0.480 \\
			POND 
			&0.885 &0.871 &0.858 &0.857 &0.463 &0.454 &0.444 &0.443 \\
			CosTCo 
			&0.999 &0.936 &0.930 &0.909 &0.523 &0.481 &0.514 &0.481 \\
			EnergyTD 
			&\underline{0.864} &\underline{0.835} &\underline{0.840} &\underline{0.833} 
			&\underline{0.450} &\underline{0.433} &\underline{0.424} &\underline{0.409} \\
			\textbf{ScoreTR}
			&\textbf{0.845} &\textbf{0.824} &\textbf{0.813} &\textbf{0.811} 
			&\textbf{0.409} &\textbf{0.397} &\textbf{0.391} &\textbf{0.387} \\
			\midrule
			\multicolumn{9}{c}{\textit{Acc} $(3k\times 150\times 30k)$}\\
			\midrule
			CP-WOPT 
			&0.533 &0.592 &0.603 &0.589 &0.138 &0.147 &0.148 &0.147 \\
			GPTF 
			&0.367 &0.357 &0.359 &0.368 &0.152 &0.150 &0.167 &0.182 \\
			HGP-GPTF 
			&0.355 &0.344 &0.341 &0.338 &0.125 &0.129 &0.139 &0.145 \\
			CosTCo 
			&0.385 &0.376 &0.363 &0.348 &0.117 &0.137 &0.107 &\underline{0.101} \\
			EnergyTD 
			&\underline{0.348} &\underline{0.336} &\underline{0.328} &\underline{0.328} 
			&\underline{0.110} &\underline{0.101} &\underline{0.094} &\underline{0.101} \\
			\textbf{ScoreTR}
			&\textbf{0.325} &\textbf{0.319} &\textbf{0.322} &\textbf{0.320} 
			&\textbf{0.076} &\textbf{0.073} &\textbf{0.073} &\textbf{0.074} \\
			\bottomrule
	\end{tabular}}
\end{table}
\begin{table}[t]
	\centering
	\small
	\caption{Quantitative results of continuous-time tensor completion.}
	\setlength{\tabcolsep}{0.6mm}
	\label{tab:continuousTC}
	\begin{tabular}{c|cccc|cccc}
		\toprule
		\multicolumn{1}{c}{Metric} &\multicolumn{4}{c}{RMSE} & \multicolumn{4}{c}{MAE} \\
		\midrule
		Method &R3 &R5 &R8 &R10 &R3 &R5 &R8 &R10 \\
		\midrule
		\multicolumn{9}{c}{\textit{Air} $(12\times 6 \times T)$}\\
		\midrule
		CTCP 
		&1.020 &1.022 &1.022 &1.022 &0.784 &0.785 &0.787 &0.787\\
		CTGP 
		&0.475 &0.463 &0.459 &0.458 &0.318 &0.304 &0.301 &0.299\\
		CTNN 
		&1.013 &1.005 &0.999 &1.013 &0.780 &0.777 &0.776 &0.780\\
		NNDTN 
		&0.377 &0.364 &0.334 &0.328 &0.247 &0.239 &0.217 &0.212\\
		NONFAT 
		&0.339 &0.335 &0.351 &0.342 &0.224 &0.219 &0.228 &0.223\\
		THIS-ODE 
		&0.569 &0.566 &0.542 &0.541 &0.415 &0.409 &0.395 &0.391\\
		EnergyTD 
		&\underline{0.302} &\underline{0.291} &\underline{0.300} &\underline{0.283} 
		&\underline{0.184} &\underline{0.177} &\underline{0.172} &\underline{0.184}\\
		\textbf{ScoreTR} 
		&\textbf{0.242} &\textbf{0.259} &\textbf{0.259} &\textbf{0.240} 
		&\textbf{0.156} &\textbf{0.170} &\textbf{0.168} &\textbf{0.154}\\
		\midrule
		\multicolumn{9}{c}{\textit{Click} $(7\times 2842\times 4127 \times T)$}\\
		\midrule
		CTCP 
		&2.063 &2.020 &2.068 &2.009 &1.000 &0.977 &1.005 &0.969\\
		CTGP 
		&1.424 &1.423 &1.404 &1.392 &0.880 &0.877 &0.856 &0.849\\
		CTNN 
		&1.820 &1.820 &1.820 &1.820 &1.077 &1.053 &1.083 & 1.071\\
		NNDTN 
		&1.418 &1.409 &1.407 &1.410 &0.858 &0.856 &0.859 &0.863\\
		NONFAT 
		&1.400 &1.411 &1.365 &1.351 &0.853 &0.873 &0.832 &0.812\\
		THIS-ODE 
		&1.421 &1.413 &1.408 &1.395 &0.836 &0.836 &0.832 &0.829\\
		EnergyTD 
		&\underline{1.396} &\underline{1.385} &\underline{1.356} &\underline{1.357} 
		&\underline{0.777} &\underline{0.775} &\underline{0.772} &\underline{0.773}\\
		\textbf{ScoreTR} 
		&\textbf{1.374} &\textbf{1.368} &\textbf{1.355} &\textbf{1.346} 
		&\textbf{0.746} &\textbf{0.749} &\textbf{0.744} &\textbf{0.747}\\
		\bottomrule
	\end{tabular}
\end{table}

\subsubsection{Sparse tensor completion}
We evaluate our model on two sparsely observed tensor datasets \cite{zhe2015scalable}: 
(1) \textit{Alog}, a file access log dataset with dimensions $200~\text{users} \times 100~\text{actions} \times 200~\text{resources}$, containing approximately 0.33\% nonzero entries; and 
(2) \textit{ACC}, a three-way tensor derived from a code repository management system with dimensions $3\text{k}~\text{users} \times 150~\text{actions} \times 30\text{k}~\text{resources}$, containing approximately 0.009\% nonzero entries. 
We follow the same data split as in \cite{tao2023undirected} and report results based on 5-fold cross-validation.

We compare our method against six baseline models: 
(1) CP-WOPT \cite{bader2008efficient}, a CP decomposition method with stochastic optimization; 
(2) GPTF \cite{zhe2016distributed}, a Gaussian process-based tensor factorization using stochastic variational inference; 
(3) HGP-GPTF \cite{tillinghast2022nonparametric}, an extension of GPTF with a hierarchical Gamma process prior; 
(4) POND \cite{tillinghast2020probabilistic}, a probabilistic non-linear TD using deep kernels with convolutional neural networks; 
(5) CoSTCo \cite{liu2019costco}, a non-linear TD that employs CNNs to map latent factors to tensor entries; and 
(6) EnergyTD \cite{tao2023undirected}, an undirected graphical TD model solved via variational conditional noise-contrastive estimation, which is the SOTA method for sparse tensor completion.
For all methods, we evaluate ranks $R \in \{3, 5, 8, 10\}$.

The completion results are presented in Table~\ref{tab:sparseTC}, where the root mean square error (RMSE) and mean absolute error (MAE) are reported as averages. Results for POND are omitted on the ACC dataset due to its high computational cost and excessive memory requirements. Our model achieves significant performance gains over the current state-of-the-art method across all cases.
The energy-based methods demonstrate notable superiority over those relying on prior contraction rules, highlighting the benefits of adopting more flexible modeling frameworks beyond the Dirac delta distribution.

\subsubsection{Continuous tensor completion}
We evaluate our model on two continuous-time tensor datasets: 
(1) \textit{Air}, the Beijing air quality dataset \cite{zhang2017cautionary}, with dimensions $12~\text{sites} \times 6~\text{pollutants}$ and approximately $1 \times 10^4$ observations across different time stamps; and 
(2) \textit{Click}, an ad click-through dataset \cite{wang2022nonparametric}, with dimensions $7~\text{banner positions} \times 2842~\text{site domains} \times 4127~\text{mobile apps}$, containing approximately $5 \times 10^4$ entries at varying time stamps. 
We follow the same data split as in \cite{wang2022nonparametric} and report results based on 5-fold cross-validation.

We compare our method with the following continuous-time tensor modeling approaches:
(1) Nonparametric factor trajectory learning (NONFAT) \cite{wang2022nonparametric}; 
(2) Continuous-time CP (CTCP) \cite{zhang2021dynamic}, which models the temporal dynamics of CP coefficients using polynomial splines; 
(3) Continuous-time GP (CTGP), an extension of GPTF \cite{zhe2016distributed} that incorporates time stamps into Gaussian process kernels; 
(4) Continuous-time NN decomposition (CTNN), a variant of CoSTCo \cite{liu2019costco} that uses time stamps as inputs to learn continuous latent trajectories; 
(5) Discrete-time NN decomposition with non-linear dynamics (NNDTN) \cite{wang2022nonparametric}, which employs RNN-based dynamics to model temporal evolution; 
(6) Tensor high-order interaction learning via ODEs (THIS-ODE) \cite{li2022decomposing}, which captures continuous-time tensor entry trajectories using neural ODEs; and 
(7) EnergyTD \cite{tao2023undirected}, an undirected graphical tensor decomposition model solved via variational conditional noise-contrastive estimation, which represents the current state-of-the-art for continuous-time tensor completion.

The completion results are presented in Table~\ref{tab:continuousTC}, with the RMSE and MAE reported as averages. Our model significantly outperforms the current state-of-the-art method across all metrics and test cases.
Most recent methods rely on minimizing squared loss under a Gaussian process assumption on the temporal data. In contrast, our model makes no such assumption, enabling more flexible adaptation to the underlying data distribution. Furthermore, by modifying the network architecture to incorporate side information, our approach improves scalability.
As an energy-based method, our model jointly learns gradients of noises at different intensities, leading to better performance than EnergyTD, which performs anti-noise training in a single stage only.

\subsection{Image Recovery}
We evaluate our model on image inpainting and denoising tasks, and compare it with state-of-the-art low-rank tensor-based methods, including: 
(1) M\textsuperscript{2}DMT \cite{fan2021multi}, a fully nonlinear framework for multi-mode deep matrix and tensor factorization; 
(2) LRTC-ENR \cite{fan2023euclideannorminduced}, which applies euclidean-norm-induced regularization based on the Schatten-$p$ quasi-norm and is solved via L-BFGS \cite{liu1989limited}; 
(3) HLRTF \cite{luo2022hlrtf}, which incorporates a DNN into the t-SVD framework using parametric total variation regularization; 
(4) DeepTensor \cite{saragadam2024deeptensor}, which represents a tensor as the product of low-rank factors generated by deep neural networks; and 
(5) LRTFR \cite{luo2023low}, which models continuous representations as low-rank tensor functions using Tucker decomposition, and is currently the state-of-the-art method for image recovery.
\begin{table}[t]
	\centering
	\small
	\caption{Average quantitative results by different methods for multispectral image denoising.}
	\label{tab:msiDenoising}
	\setlength{\tabcolsep}{1mm} 
	\begin{tabular}{c|ccc|ccc}
		\toprule
		\multicolumn{7}{c}{MSIs \textit{Balloons, Beads, Flowers, Fruits }$(512\times 512\times 31)$} \\
		\midrule
		\multicolumn{1}{c|}{Method} & PSNR & SSIM & NRMSE & PSNR & SSIM & NRMSE \\
		\midrule
		\multicolumn{1}{c}{Noise} & \multicolumn{3}{c}{Case 1} & \multicolumn{3}{c}{Case 2} \\
		\midrule
		\textit{Observed} 
		&16.22 & 0.084 & 0.902 & 16.26 & 0.109 & 0.900 \\
		M\textsuperscript{2}DMT
		&29.80 & 0.720 & 0.158 & 30.94 & 0.748 & 0.136 \\
		LRTC-ENR
		&31.26 & \underline{0.756} & \underline{0.152} & \underline{33.88} & \underline{0.845} & \underline{0.127} \\
		HLRTF
		&30.57 & 0.731 & 0.159 & 32.75 & 0.781 & 0.152 \\
		DeepTensor
		& 29.97 & 0.725 & 0.155 & 31.03 & 0.747 & 0.140 \\
		LRTFR
		&\underline{31.32} & 0.736 & 0.167 & 32.89 & 0.784 & 0.141 \\
		\textbf{ScoreTR} 
		&\textbf{32.32} & \textbf{0.856} & \textbf{0.150} & \textbf{34.54} & \textbf{0.865} & \textbf{0.118} \\
		\midrule
		\multicolumn{1}{c}{Noise} & \multicolumn{3}{c}{Case 3} & \multicolumn{3}{c}{Case 4} \\
		\midrule
		\textit{Observed} 
		&16.12 & 0.101 & 0.912 & 16.20 & 0.107 & 0.906 \\
		M\textsuperscript{2}DMT
		&30.91 & 0.747 & 0.170 & 28.32 & 0.731 & 0.179 \\
		LRTC-ENR
		&31.49 & 0.770 & 0.156 & 28.96 & 0.753 & 0.174 \\
		HLRTF
		&31.51 & 0.791 & 0.158 & 29.53 & 0.756 & 0.173 \\
		DeepTensor
		&30.79 & 0.786 & 0.169 & 29.77 & 0.741 & 0.179 \\
		LRTFR
		&\underline{31.96} & \underline{0.794} & \underline{0.153} & \underline{31.27} & \underline{0.776} & \underline{0.162} \\
		\textbf{ScoreTR} 
		&\textbf{32.51} & \textbf{0.870} & \textbf{0.144} & \textbf{31.83} & \textbf{0.835} & \textbf{0.154} \\
		\midrule
		\multicolumn{1}{c}{Noise} & \multicolumn{3}{c|}{Case 5} & \multicolumn{3}{c}{Case 6} \\
		\midrule
		\textit{Observed} 
		&16.07 & 0.101 & 0.917 & 17.79 & 0.290 & 0.760 \\
		M\textsuperscript{2}DMT
		&27.17 & 0.769 & 0.230 & 35.72 & 0.944 & 0.103 \\
		LRTC-ENR
		&28.07 & 0.780 & 0.216 & \underline{39.66} & 0.968 & 0.076 \\
		HLRTF
		&27.93 & 0.774 & 0.228 & 37.81 & 0.967 & 0.081 \\
		DeepTensor
		&27.89 & 0.772 & 0.221 & 37.19 & 0.963 & 0.087 \\
		LRTFR
		&\textbf{29.97} & \underline{0.782} & \textbf{0.187} & 39.32 & \underline{0.971} & \underline{0.068} \\
		\textbf{ScoreTR} 
		&\underline{29.71} & \textbf{0.841} & \underline{0.195} & \textbf{43.83} & \textbf{0.988} & \textbf{0.042} \\
		\bottomrule
	\end{tabular}
\end{table}
\begin{table*}[t]
	\centering
	\small
	\caption{Average quantitative results of multidimensional images by different methods.}
	\label{tab:imageInpainting}
	\begin{tabular}{cc|ccc|ccc|ccc}
		\toprule
		\multicolumn{2}{c}{Sampling rate} & \multicolumn{3}{c}{0.1} & \multicolumn{3}{c}{0.15} & \multicolumn{3}{c}{0.2} \\
		\midrule
		Data &Method &PSNR &SSIM &NRMSE &PSNR &SSIM &NRMSE &PSNR &SSIM &NRMSE\\
		\midrule
		\multirow{7}{*}{\makecell[c]{Color images\\\textit{Sailboat}\\\textit{House}\\\textit{Peppers}\\\textit{Plane}\\$(512\times 512\times 3)$}} 
		&\textit{Observed} 
		&4.846 &0.023 &0.949 &5.095 &0.030 &0.922 &5.358 &0.038 &0.895 \\
		&M\textsuperscript{2}DMT 
		&22.06 &0.573 &0.145 
		&23.49 &0.650 &0.137 
		&23.89 &0.692 &0.116 \\
		&LRTC-ENR 
		&\underline{23.56} &\underline{0.628} &\underline{0.128} 
		&24.61 &0.694 &0.114 
		&25.16 &0.707 &0.102 \\
		&HLRTF 
		&22.49 &0.540 &0.136 
		&24.41 &0.679 &0.110 
		&25.39 &0.711 &0.097\\
		&DeepTensor 
		&21.50 &0.484 &0.150 
		&24.53 &0.682 &0.118 
		&26.31 &0.717 &0.101 \\
		&LRTFR 
		&23.03 &0.597 &0.132 
		&\underline{26.22} &\underline{0.695} &\underline{0.084} 
		&\underline{27.49} &\underline{0.741} &\underline{0.073}\\
		&\textbf{ScoreTR} 
		&\textbf{25.34} &\textbf{0.755} &\textbf{0.092} 
		&\textbf{27.22} &\textbf{0.804} &\textbf{0.075} 
		&\textbf{28.14} &\textbf{0.829} &\textbf{0.068} \\
		\midrule
		\multirow{7}{*}{\makecell[c]{MSIs\\\textit{Toys}\\\textit{Flowers}\\$(512\times 512\times 31)$}}
		&\textit{Observed} 
		&13.96 &0.386 &0.949 
		&14.21 &0.418 &0.922 
		&14.47 &0.447 &0.894\\
		&M\textsuperscript{2}DMT 
		&34.89 &0.910 &0.107 &36.82 &0.928 &0.092 
		&38.19 &0.934 &0.082\\
		&LRTC-ENR 
		&35.91 &0.928 &0.094 
		&37.14 &0.935 &0.080 
		&39.33 &0.950 &0.070\\
		&HLRTF 
		&36.32 &0.935 &0.091 
		&38.64 &0.942 &0.076 
		&40.19 &0.955 &0.067\\
		&DeepTensor 
		&38.40 &0.947 &0.088 
		&39.99 &0.951 &0.077 
		&41.20 &0.965 &0.066\\
		&LRTFR 
		&\underline{40.16} &\underline{0.969} &\underline{0.047} 
		&\underline{42.74} &\underline{0.982} &\underline{0.035} 
		&\underline{44.28} &\underline{0.985} &\underline{0.029}\\
		&\textbf{ScoreTR} 
		&\textbf{41.27} &\textbf{0.985} &\textbf{0.042}
		&\textbf{44.22} &\textbf{0.990} &\textbf{0.030}
		&\textbf{46.28} &\textbf{0.993} &\textbf{0.024}\\
		\midrule
		\multirow{7}{*}{\makecell[c]{Videos\\\textit{Foreman}\\\textit{Carphone}\\$(144\times 176\times 100)$}}
		&\textit{Observed} 
		&5.548 &0.017 &0.949 &5.797 &0.024 &0.922 &6.059 &0.031 &0.894\\
		&M\textsuperscript{2}DMT 
		&23.51 &0.701 &0.124 &25.21 &0.769 &0.102 &26.47 &0.815 &0.095\\
		&LRTC-ENR 
		&24.23 &0.730 &0.117 &25.91 &0.793 &0.094 &27.56 &0.826 &0.088\\
		&HLRTF 
		&24.66 &0.768 &0.104 &26.49 &0.830 &0.085 &28.10 &0.837 &0.071\\
		&DeepTensor 
		&25.67 &0.813 &0.114 &27.34 &0.855 &0.080 &28.89 &0.851 &0.067\\
		&LRTFR 
		&\underline{28.53} &\underline{0.828} &\underline{0.067} 
		&\underline{29.36} &\underline{0.854} &\underline{0.061} 
		&\underline{29.77} &\underline{0.866} &\underline{0.058}\\
		&\textbf{ScoreTR} 
		&\textbf{33.35} &\textbf{0.945} &\textbf{0.039} 
		&\textbf{35.04} &\textbf{0.958} &\textbf{0.032} 
		&\textbf{36.30} &\textbf{0.966} &\textbf{0.028}\\
		\bottomrule
	\end{tabular}
\end{table*}

\subsubsection{Image Denoising}
MSI denoising aims to recover a clean image from noisy observations. In practice, MSIs are often degraded by mixed noise types, including Gaussian noise, sparse noise, stripe noise, and dead lines.
The experiments are conducted on four MSIs from the CAVE dataset\footnote{\label{foot:cave}\url{https://www.cs.columbia.edu/CAVE/databases/multispectral/}}. According to the experimental setup of LRTFR\cite{luo2023low}, we consider six noise scenarios to comprehensively evaluate the robustness and effectiveness of denoising algorithms under various noise conditions. (1) Case 1 introduces Gaussian noise with a standard deviation of 0.2; (2) Case 2 combines Gaussian noise ($\sigma = 0.1$) with sparse noise at a sparsity rate of 0.1; (3) Case 3 extends Case 2 by adding dead lines across all spectral bands; (4) Case 4 includes the same noise as in Case 2 along with 10\% stripe noise in 40\% of the spectral bands; (5) Case 5 further incorporates dead lines into Case 4, combining stripe noise and dead lines for a more challenging scenario; and (6) Case 6 considers only sparse noise with rate of 0.1.
These scenarios provide a comprehensive benchmark for evaluating MSI denoising methods under diverse and realistic noise conditions.

The completion results are presented in Table~\ref{tab:msiDenoising}, where the average values of PSNR, SSIM, and NRMSE are reported. Our model significantly outperforms the current state-of-the-art method in most cases, particularly in terms of SSIM.
We incorporate the proposed smooth loss from an energy perspective into the block coordinate descent framework, enabling effective handling of the MSI denoising task.
It has been shown that assigning lower energy to tensor values and their neighboring factors can achieve effective smooth regularization. This approach avoids the computational burden of TV regularization, which requires full tensor posterior sampling in each iteration and is therefore time-consuming.

\subsubsection{Image Inpainting}
We evaluate our model on three type of image data: (1) \textit{RGB image}\footnote{\url{https://sipi.usc.edu/database/database.php}}; (2) \textit{Multispectral image}\footref{foot:cave}; and (3) \textit{Video}\footnote{\url{http://trace.eas.asu.edu/yuv/}}. We evaluated performance under random missing conditions with sampling rates (SRs) of 10\%, 15\%, 20\%.

The completion results are presented in Table~\ref{tab:imageInpainting}, where the peak signal-to-noise ratio (PSNR), structural similarity index (SSIM), and normalized root mean square error (NRMSE) are reported as averages. Our model significantly outperforms the current state-of-the-art method across all test cases, with particularly notable improvements on video data, demonstrating its strong performance on visual information.
The superior performance of our method can be attributed to its ability to jointly encode low-rankness and smoothness into the learned representation.
Traditional methods heavily rely on predefined contraction structures and rank definitions, making them less adaptable to tensors from diverse domains, as the intrinsic structures of such tensors often vary significantly. In contrast, our approach is capable of handling tensor data from multiple fields simultaneously, without being constrained by prior contraction rules.

\section{Conclusion}
We introduce an innovative low-rank tensor recovery model solved via denoising score matching, distinguished by its flexibility in adapting to diverse structures and distributions. Our method eliminates the need for computing expensive high-dimensional tensor contractions, thereby overcoming the limitations of traditional approaches that rely on predefined contraction rules. By learning the gradient of the joint probability distribution of tensor entries and latent factors, it achieves superior performance over the state-of-the-art methods across multiple domains and tasks.


\clearpage
\bibliography{aaai2026}

\newpage
\onecolumn
\appendix
\section{Experimental settings}
All experiments were performed on a system equipped with two Intel Xeon Silver 4214 CPU and NVIDIA RTX 3090 GPUs.
The specific implementation details of our experiments are summarized in Table~\ref{tab:implementation}. Due to differences in data size and value range across datasets, we adopted distinct training parameters for each experiment. For each dataset, we employ a two-layer MLP with uniform width to parameterize all instances of $g_\theta(\cdot)$, and the corresponding network width is given in Column 6 of Table~\ref{tab:implementation}.
\begin{table*}[h]
	\centering
	\small
	\caption{The implementation details of the experiments.}
	\label{tab:implementation}
	\begin{tabular}{cccccccc}
		\toprule
		Task &Dataset &Epoch &Batch &$\sigma_{max}, \sigma_{min}, L$ &Network &Ir &Rank \\
		\midrule
		\multirow{2}{*}{Sparse tensor completion}
		&Alog &1000 &256 &0.2, 0.01, 10 &[256, 256] &1e-3 &\{3,5,8,10\} \\
		&Acc &100 &8192 &0.1, 0.01, 10 &[512, 512] &1e-3 &\{3,5,8,10\} \\
		\midrule
		\multirow{2}{*}{Continuous tensor completion}
		&Air &1000 &128  &1.0, 0.01, 10 &[50, 50] &5e-3 &\{3,5,8,10\} \\
		&Click &200 &128  &1.0, 0.01, 10 &[64, 64] &1e-3 &\{3,5,8,10\} \\
		\midrule
		\multirow{3}{*}{Image inpainting}
		&RGB &400 &8192  &0.1, 0.01, 10 &[512, 512] &1e-3 &256 \\
		&MSI &400 &8192  &0.1, 0.001, 10 &[384, 384] &1e-3 &256 \\
		&Video &400 &8192 &0.1, 0.01, 10 &[512, 512] &1e-3 &256 \\
		\midrule
		Image denoising &MSI &100 &8192 &0.1, 0.001, 10 &[384, 384] &1e-3 &128\\
		\bottomrule
	\end{tabular}
\end{table*}

\section{Discussion}
Selecting appropriate hyperparameter values is a critical step in our method, particularly the choice of $\sigma_{\max}$ and the number of noise levels $L$. 
We conducted experiments on all types of tensors in data recovery, the ablation studies indicate that these hyperparameters have varying degrees of influence on overall performance. In most cases, a lower Gaussian noise variance yields better results. Our analysis further shows that the gradient of the learned energy landscape becomes more refined as the noise variance decreases, which supports the effectiveness of the proposed method.
\begin{figure*}[h]
	\centering
	\captionsetup[subfloat]{labelsep=none,format=plain,labelformat=empty}
	\begin{minipage}{1.\linewidth}
		\centering
		\subfloat[]
		{\includegraphics[width=0.12\linewidth]{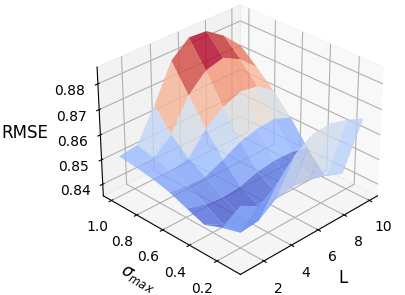}}
		\subfloat[]
		{\includegraphics[width=0.12\linewidth]{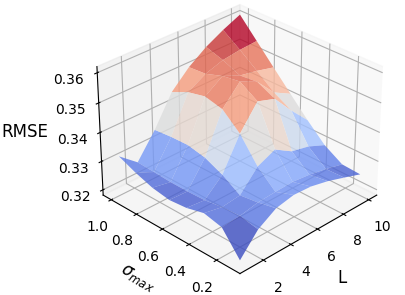}}
		\subfloat[]
		{\includegraphics[width=0.12\linewidth]{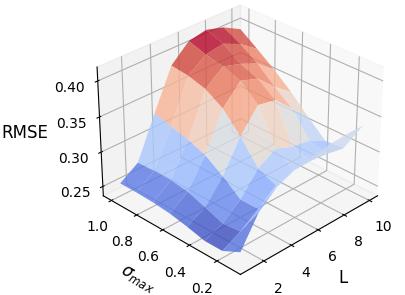}}
		\subfloat[]
		{\includegraphics[width=0.12\linewidth]{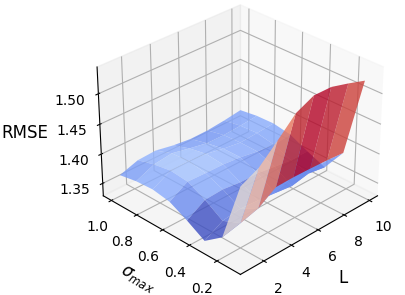}}
		\subfloat[]
		{\includegraphics[width=0.12\linewidth]{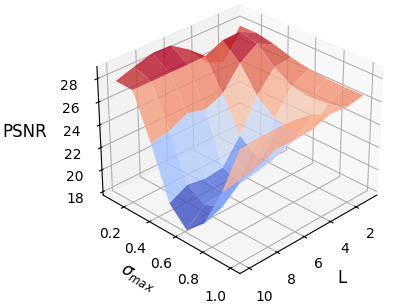}}
		\subfloat[]
		{\includegraphics[width=0.12\linewidth]{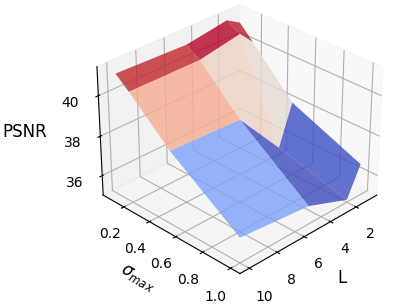}}
		\subfloat[]
		{\includegraphics[width=0.12\linewidth]{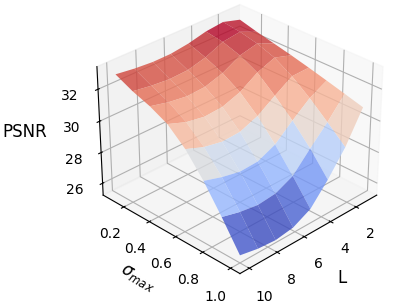}}
	\end{minipage}
	\begin{minipage}{1.\linewidth}
		\centering
		\subfloat[Alog]
		{\includegraphics[width=0.12\linewidth]{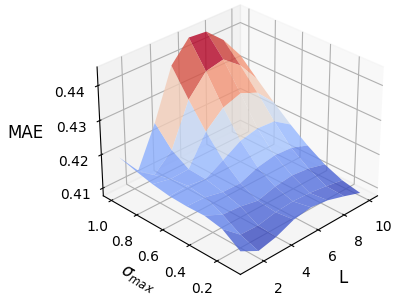}}
		\subfloat[Acc]
		{\includegraphics[width=0.12\linewidth]{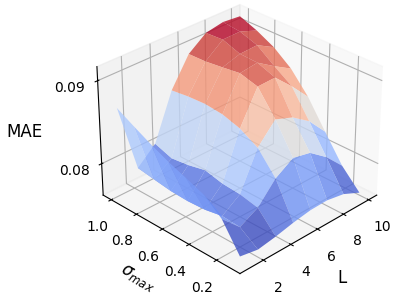}}
		\subfloat[Air]
		{\includegraphics[width=0.12\linewidth]{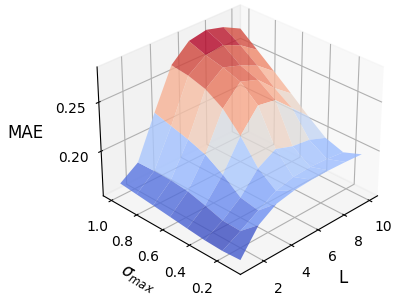}}
		\subfloat[Click]
		{\includegraphics[width=0.12\linewidth]{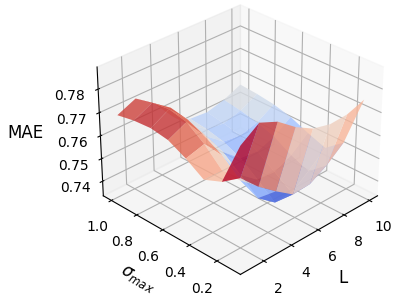}}
		\subfloat[RGB]
		{\includegraphics[width=0.12\linewidth]{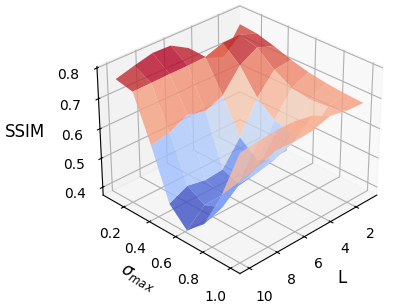}}
		\subfloat[MSI]
		{\includegraphics[width=0.12\linewidth]{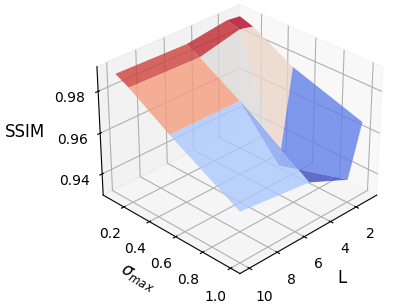}}
		\subfloat[Video]
		{\includegraphics[width=0.12\linewidth]{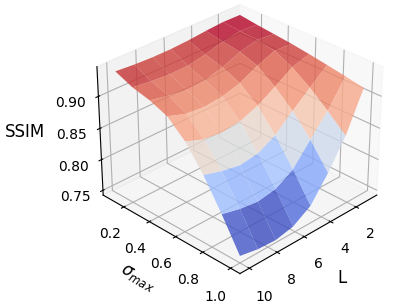}}
	\end{minipage}
	\caption{Effect of $\sigma_{max}$ and the number of noise level $L$ on spare different tensors.}
	\label{fig:ablation1}
\end{figure*}

\section{Visualization}
We present a visual comparison of the results for image completion and denoising tasks. 
\begin{figure*}[h]
	\centering
	\captionsetup[subfloat]{labelsep=none,format=plain,labelformat=empty}
	\begin{minipage}{1.\linewidth}
		\centering
		\subfloat[PSNR 5.609]
		{\includegraphics[width=0.115\linewidth]{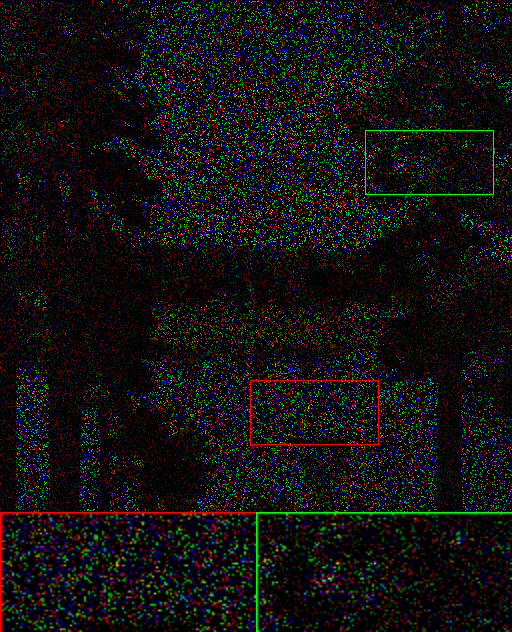}}
		\hspace{0.01cm}
		\subfloat[PSNR 19.61]
		{\includegraphics[width=0.115\linewidth]{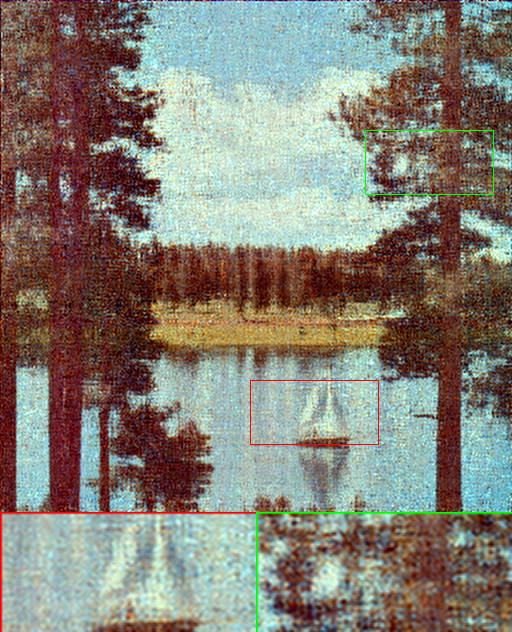}}
		\hspace{0.01cm}
		\subfloat[PSNR 20.40]
		{\includegraphics[width=0.115\linewidth]{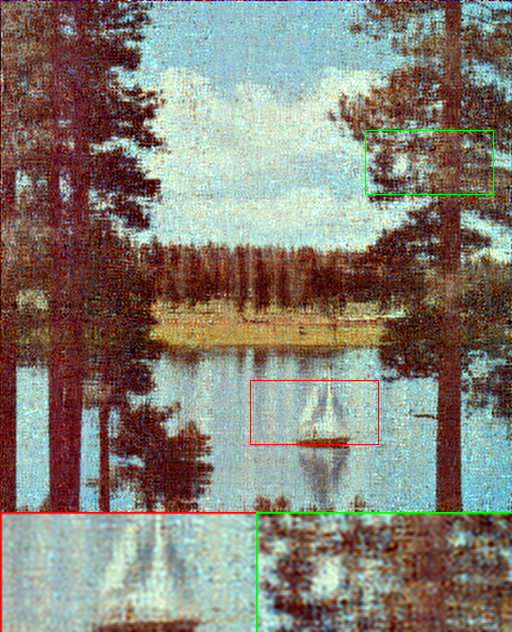}}
		\hspace{0.01cm}
		\subfloat[PSNR 21.21]
		{\includegraphics[width=0.115\linewidth]{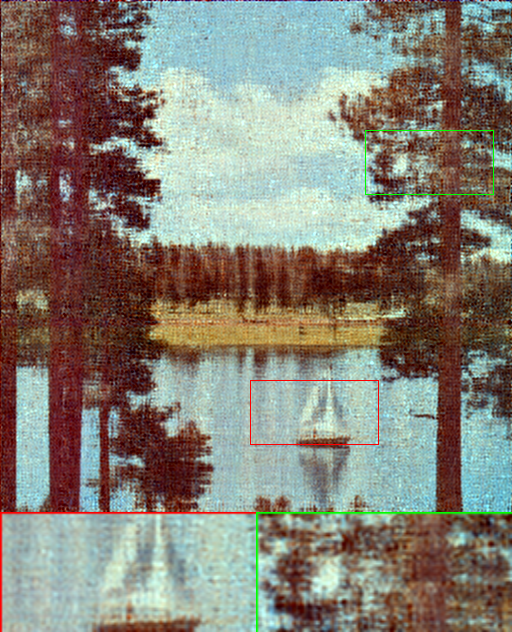}}
		\hspace{0.01cm}
		\subfloat[PSNR 21.63]
		{\includegraphics[width=0.115\linewidth]{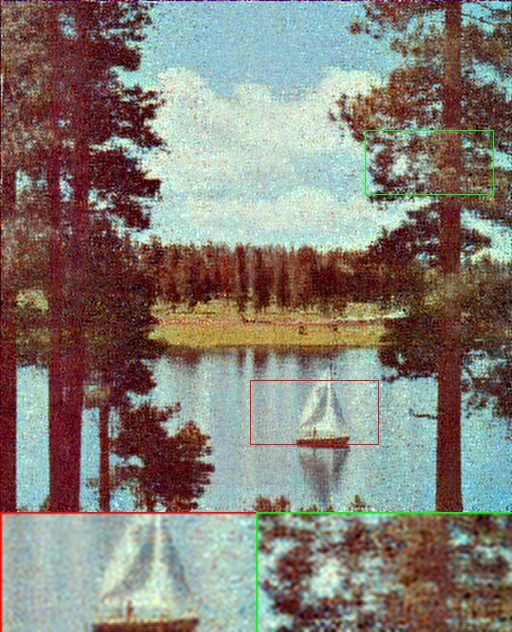}}
		\hspace{0.01cm}
		\subfloat[PSNR 21.86]
		{\includegraphics[width=0.115\linewidth]{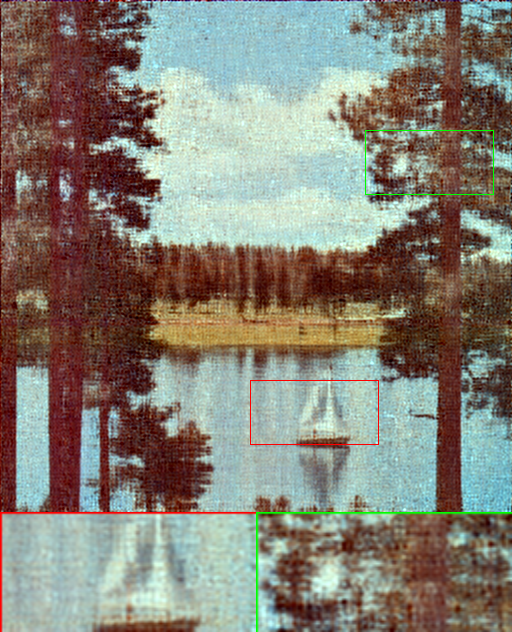}}
		\hspace{0.01cm}
		\subfloat[PSNR 23.52]
		{\includegraphics[width=0.115\linewidth]{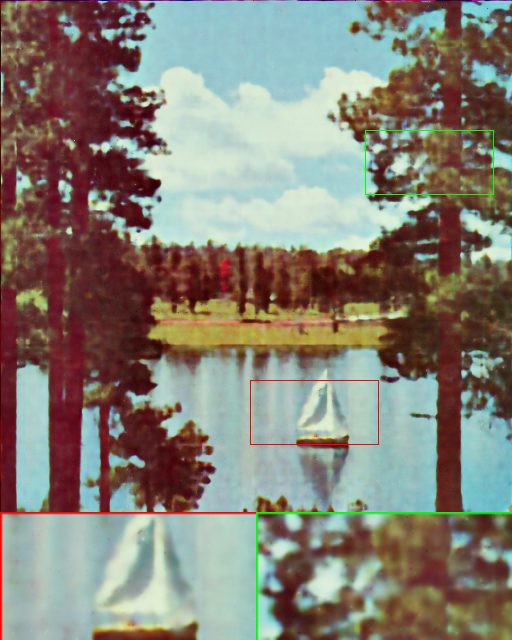}}
		\hspace{0.01cm}
		\subfloat[PSNR Inf]
		{\includegraphics[width=0.115\linewidth]{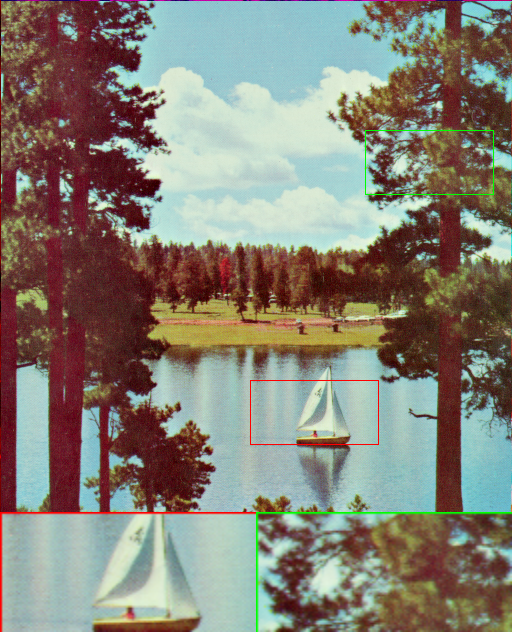}}
	\end{minipage}
	\begin{minipage}{1.\linewidth}
		\centering
		\subfloat[PSNR 6.892]
		{\includegraphics[width=0.115\linewidth]{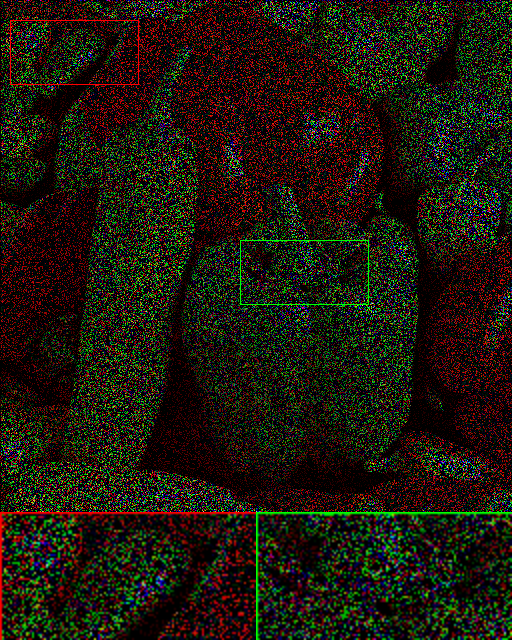}}
		\hspace{0.01cm}
		\subfloat[PSNR 23.70]
		{\includegraphics[width=0.115\linewidth]{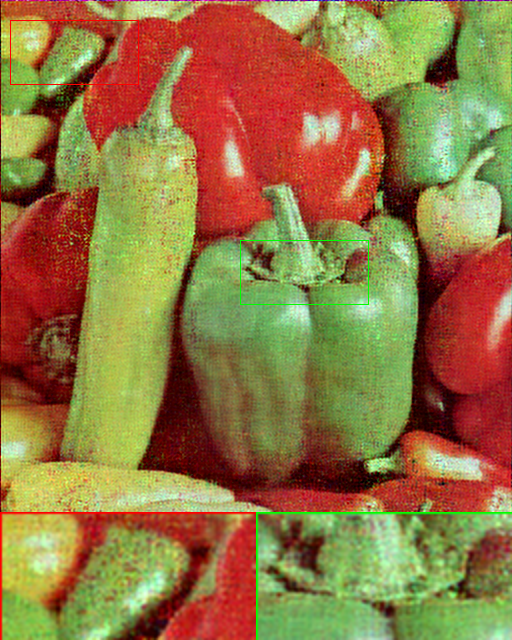}}
		\hspace{0.01cm}
		\subfloat[PSNR 26.65]
		{\includegraphics[width=0.115\linewidth]{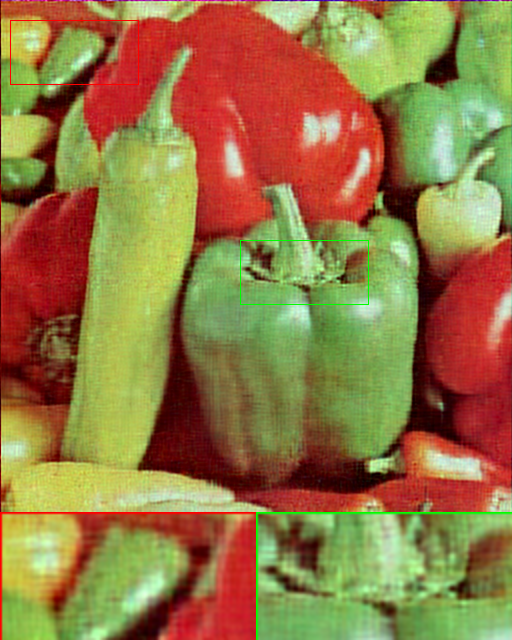}}
		\hspace{0.01cm}
		\subfloat[PSNR 26.93]
		{\includegraphics[width=0.115\linewidth]{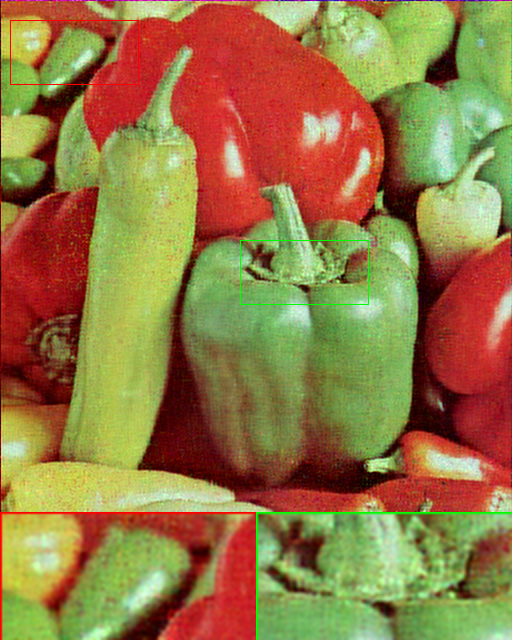}}
		\hspace{0.01cm}
		\subfloat[PSNR 25.96]
		{\includegraphics[width=0.115\linewidth]{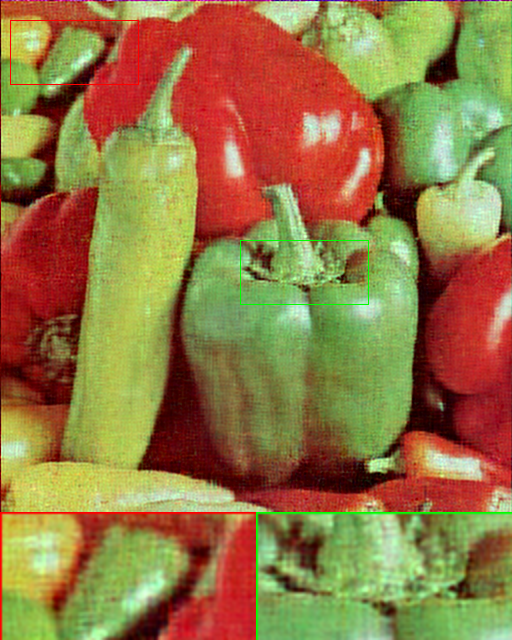}}
		\hspace{0.01cm}
		\subfloat[PSNR 28.24]
		{\includegraphics[width=0.115\linewidth]{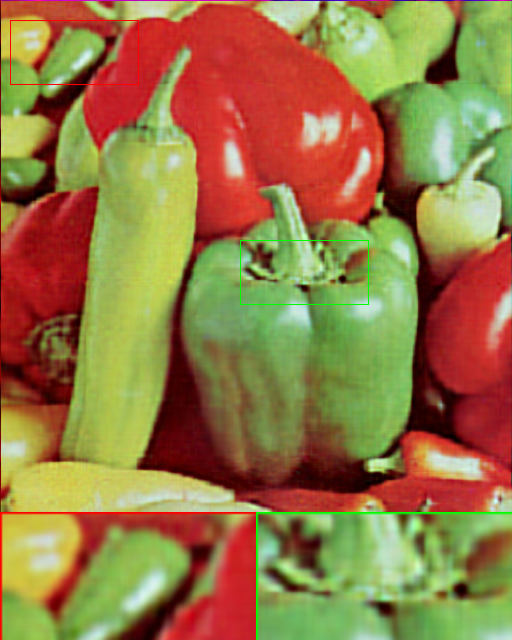}}
		\hspace{0.01cm}
		\subfloat[PSNR 30.09]
		{\includegraphics[width=0.115\linewidth]{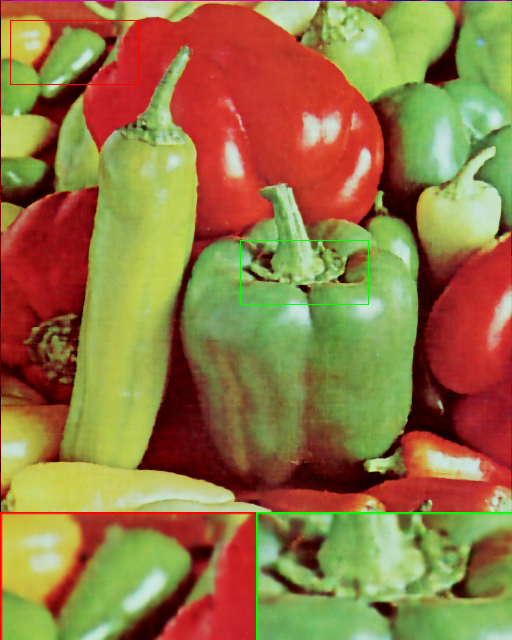}}
		\hspace{0.01cm}
		\subfloat[PSNR Inf]
		{\includegraphics[width=0.115\linewidth]{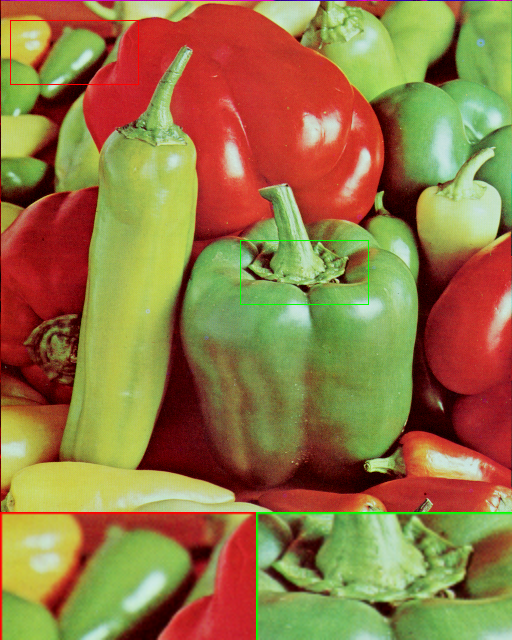}}
	\end{minipage}
	\begin{minipage}{1.\linewidth}
		\centering
		\subfloat[PSNR 11.49]
		{\includegraphics[width=0.115\linewidth]{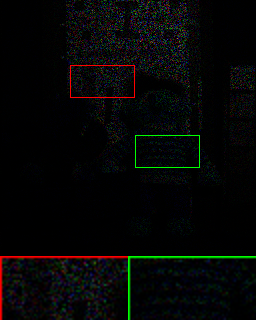}}
		\hspace{0.01cm}
		\subfloat[PSNR 34.59]
		{\includegraphics[width=0.115\linewidth]{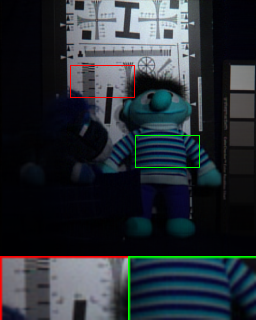}}
		\hspace{0.01cm}
		\subfloat[PSNR 36.87]
		{\includegraphics[width=0.115\linewidth]{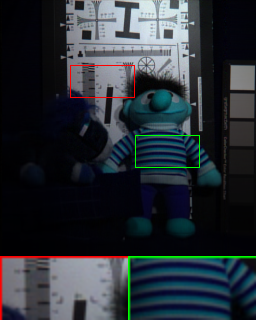}}
		\hspace{0.01cm}
		\subfloat[PSNR 38.24]
		{\includegraphics[width=0.115\linewidth]{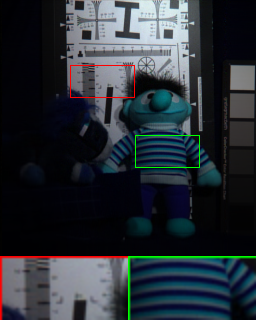}}
		\hspace{0.01cm}
		\subfloat[PSNR 39.17]
		{\includegraphics[width=0.115\linewidth]{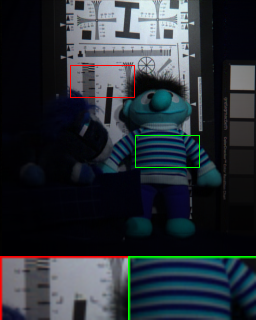}}
		\hspace{0.01cm}
		\subfloat[PSNR 40.39]
		{\includegraphics[width=0.115\linewidth]{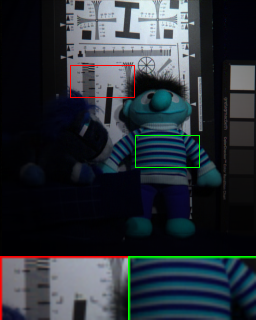}}
		\hspace{0.01cm}
		\subfloat[PSNR 41.07]
		{\includegraphics[width=0.115\linewidth]{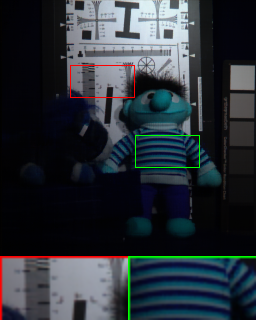}}
		\hspace{0.01cm}
		\subfloat[PSNR Inf]
		{\includegraphics[width=0.115\linewidth]{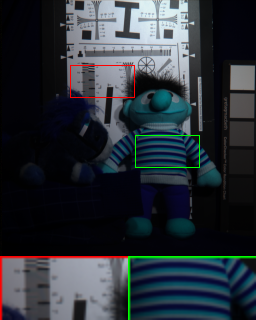}}
	\end{minipage}
	\begin{minipage}{1.\linewidth}
		\centering
		\subfloat[PSNR 16.94]
		{\includegraphics[width=0.115\linewidth]{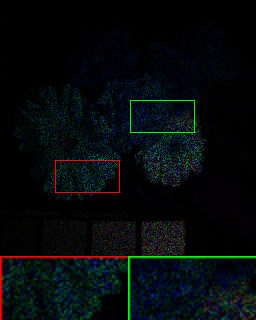}}
		\hspace{0.01cm}
		\subfloat[PSNR 39.52]
		{\includegraphics[width=0.115\linewidth]{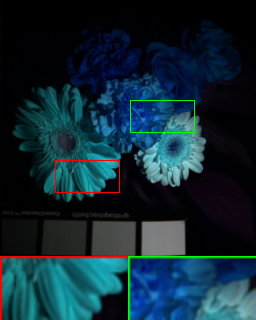}}
		\hspace{0.01cm}
		\subfloat[PSNR 41.00]
		{\includegraphics[width=0.115\linewidth]{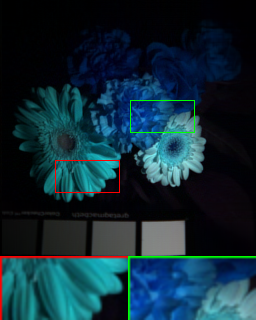}}
		\hspace{0.01cm}
		\subfloat[PSNR 44.08]
		{\includegraphics[width=0.115\linewidth]{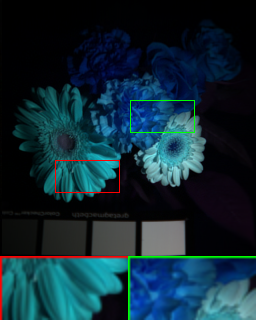}}
		\hspace{0.01cm}
		\subfloat[PSNR 42.36]
		{\includegraphics[width=0.115\linewidth]{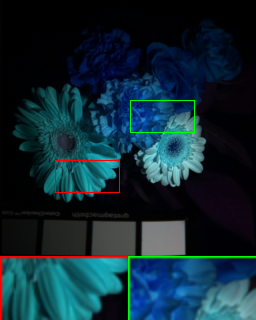}}
		\hspace{0.01cm}
		\subfloat[PSNR 45.30]
		{\includegraphics[width=0.115\linewidth]{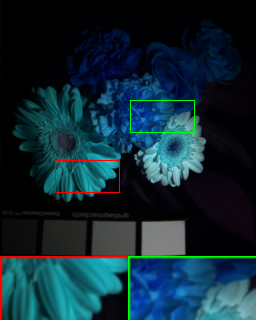}}
		\hspace{0.01cm}
		\subfloat[PSNR 46.07]
		{\includegraphics[width=0.115\linewidth]{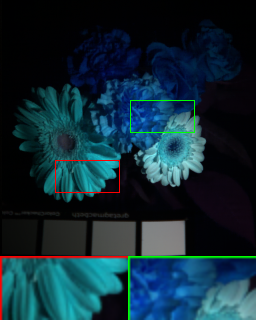}}
		\hspace{0.01cm}
		\subfloat[PSNR Inf]
		{\includegraphics[width=0.115\linewidth]{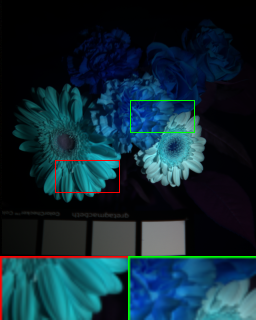}}
	\end{minipage}
	\begin{minipage}{1.\linewidth}
		\centering
		\subfloat[PSNR 4.048]
		{\includegraphics[width=0.115\linewidth]{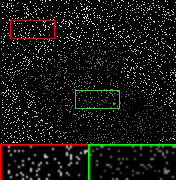}}
		\hspace{0.01cm}
		\subfloat[PSNR 24.46]
		{\includegraphics[width=0.115\linewidth]{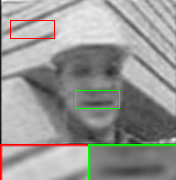}}
		\hspace{0.01cm}
		\subfloat[PSNR 25.60]
		{\includegraphics[width=0.115\linewidth]{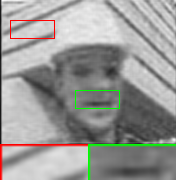}}
		\hspace{0.01cm}
		\subfloat[PSNR 25.91]
		{\includegraphics[width=0.115\linewidth]{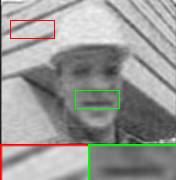}}
		\hspace{0.01cm}
		\subfloat[PSNR 27.05]
		{\includegraphics[width=0.115\linewidth]{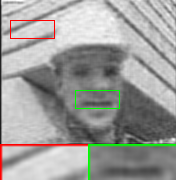}}
		\hspace{0.01cm}
		\subfloat[PSNR 27.14]
		{\includegraphics[width=0.115\linewidth]{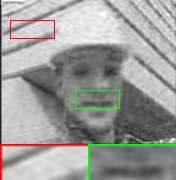}}
		\hspace{0.01cm}
		\subfloat[PSNR 32.92]
		{\includegraphics[width=0.115\linewidth]{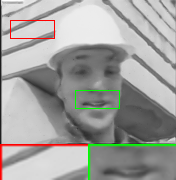}}
		\hspace{0.01cm}
		\subfloat[PSNR Inf]
		{\includegraphics[width=0.115\linewidth]{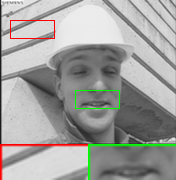}}
	\end{minipage}
	\begin{minipage}{1.\linewidth}
		\centering
		\subfloat[\begin{tabular}{c} PSNR 7.559\\ Observed\end{tabular}]
		{\includegraphics[width=0.115\linewidth]{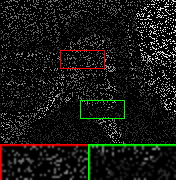}}
		\hspace{0.01cm}
		\subfloat[\begin{tabular}{c} PSNR 27.25\\ M\textsuperscript{2}DMT \end{tabular}]
		{\includegraphics[width=0.115\linewidth]{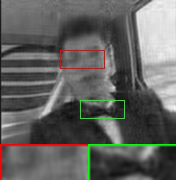}}
		\hspace{0.01cm}
		\subfloat[\begin{tabular}{c} PSNR 27.42\\ LRTC-ENR \end{tabular}]
		{\includegraphics[width=0.115\linewidth]{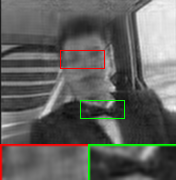}}
		\hspace{0.01cm}
		\subfloat[\begin{tabular}{c}PSNR 28.85\\ HLRTF \end{tabular}]
		{\includegraphics[width=0.115\linewidth]{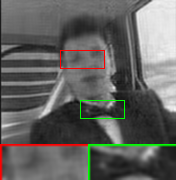}}
		\hspace{0.01cm}
		\subfloat[\begin{tabular}{c}PSNR 29.81\\ DeepTensor \end{tabular}]
		{\includegraphics[width=0.115\linewidth]{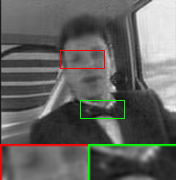}}
		\hspace{0.01cm}
		\subfloat[\begin{tabular}{c}PSNR 31.31\\ LRTFR \end{tabular}]
		{\includegraphics[width=0.115\linewidth]{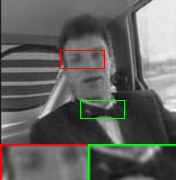}}
		\hspace{0.01cm}
		\subfloat[\begin{tabular}{c}PSNR 36.41\\ ScoreTR \end{tabular}]
		{\includegraphics[width=0.115\linewidth]{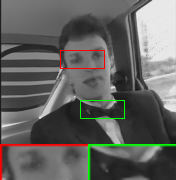}}
		\hspace{0.01cm}
		\subfloat[\begin{tabular}{c}PSNR Inf\\ Original \end{tabular}]
		{\includegraphics[width=0.115\linewidth]{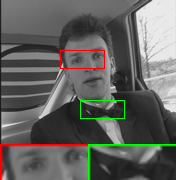}}
	\end{minipage}
	\caption{Results of multi-dimensional image inpainting by different methods on the color images \textit{Sailboat} and \textit{Peppers}, multispectral images \textit{Toys} and \textit{Flowers}, and videos \textit{Foreman} and \textit{Carphone}. The sampling rates for rows 1, 3, and 5 are 0.1, and those for rows 2, 4, and 6 are 0.2.}
	\label{fig:Demo4Inpainting}
\end{figure*}
The qualitative results of multi-dimensional image inpainting are shown in Figure~\ref{fig:Demo4Inpainting}. It can be observed that our ScoreTR achieves the best performance qualitatively, demonstrating its superiority over traditional low-rank tensor representations. Notably, the improvement is especially evident in the video data, where the reconstructed images are visually clearer and more natural compared to previous methods.

\begin{figure*}[h]
	\centering
	\captionsetup[subfloat]{labelsep=none,format=plain,labelformat=empty}
	\begin{minipage}{1.\linewidth}
		\centering
		\subfloat[PSNR 15.91]
		{\includegraphics[width=0.115\linewidth]{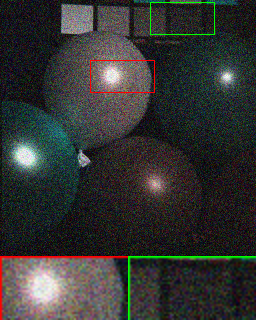}}
		\hspace{0.01cm}
		\subfloat[PSNR 33.13]
		{\includegraphics[width=0.115\linewidth]{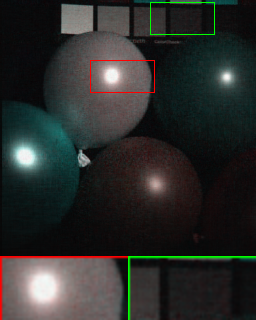}}
		\hspace{0.01cm}
		\subfloat[PSNR 36.11]
		{\includegraphics[width=0.115\linewidth]{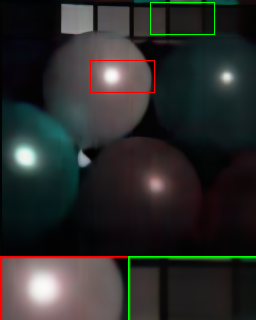}}
		\hspace{0.01cm}
		\subfloat[PSNR 35.45]
		{\includegraphics[width=0.115\linewidth]{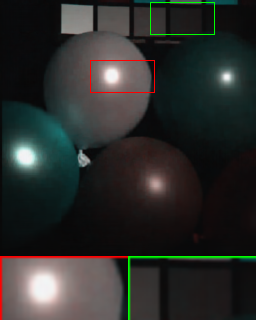}}
		\hspace{0.01cm}
		\subfloat[PSNR 35.19]
		{\includegraphics[width=0.115\linewidth]{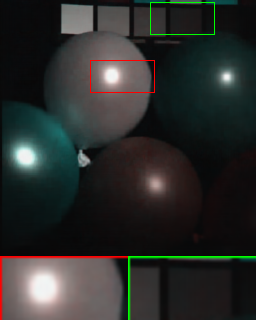}}
		\hspace{0.01cm}
		\subfloat[PSNR 36.62]
		{\includegraphics[width=0.115\linewidth]{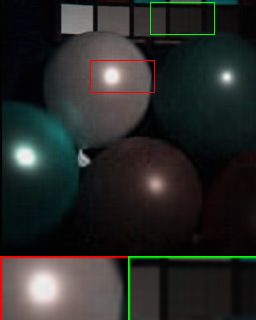}}
		\hspace{0.01cm}
		\subfloat[PSNR 37.73]
		{\includegraphics[width=0.115\linewidth]{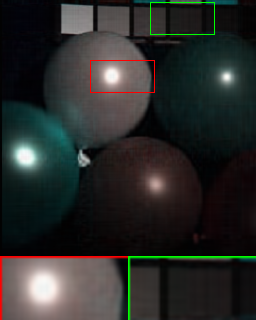}}
		\hspace{0.01cm}
		\subfloat[PSNR Inf]
		{\includegraphics[width=0.115\linewidth]{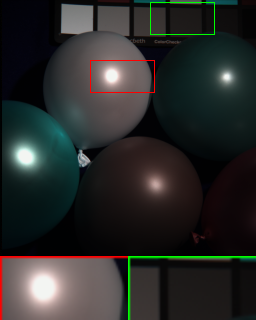}}
	\end{minipage}
	\begin{minipage}{1.\linewidth}
		\centering
		\subfloat[PSNR 16.30]
		{\includegraphics[width=0.115\linewidth]{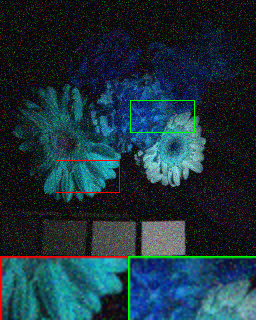}}
		\hspace{0.01cm}
		\subfloat[PSNR 32.68]
		{\includegraphics[width=0.115\linewidth]{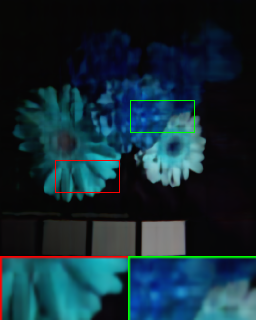}}
		\hspace{0.01cm}
		\subfloat[PSNR 36.17]
		{\includegraphics[width=0.115\linewidth]{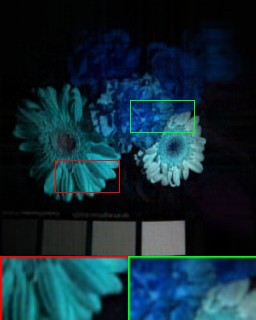}}
		\hspace{0.01cm}
		\subfloat[PSNR 34.49]
		{\includegraphics[width=0.115\linewidth]{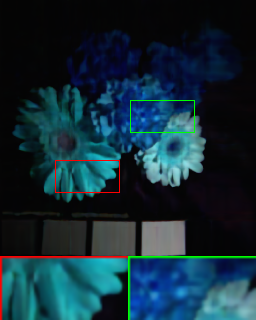}}
		\hspace{0.01cm}
		\subfloat[PSNR 34.07]
		{\includegraphics[width=0.115\linewidth]{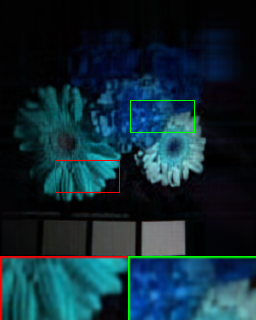}}
		\hspace{0.01cm}
		\subfloat[PSNR 35.20]
		{\includegraphics[width=0.115\linewidth]{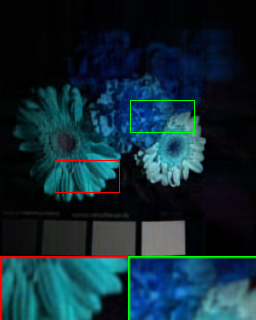}}
		\hspace{0.01cm}
		\subfloat[PSNR 37.31]
		{\includegraphics[width=0.115\linewidth]{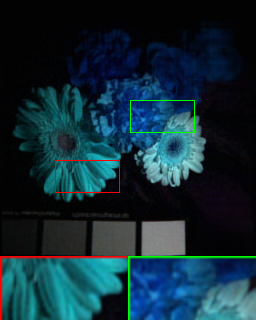}}
		\hspace{0.01cm}
		\subfloat[PSNR Inf]
		{\includegraphics[width=0.115\linewidth]{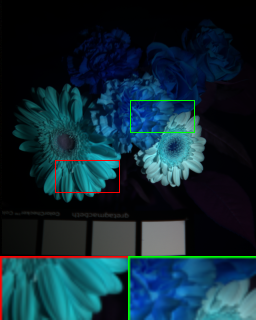}}
	\end{minipage}
	\begin{minipage}{1.\linewidth}
		\centering
		\subfloat[PSNR 15.94]
		{\includegraphics[width=0.115\linewidth]{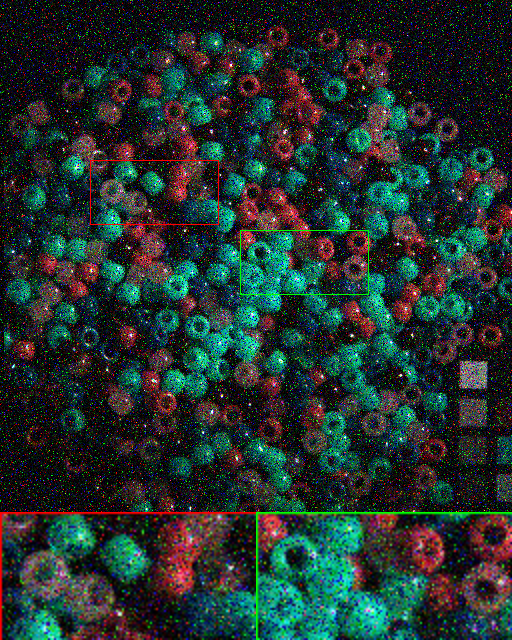}}
		\hspace{0.01cm}
		\subfloat[PSNR 26.68]
		{\includegraphics[width=0.115\linewidth]{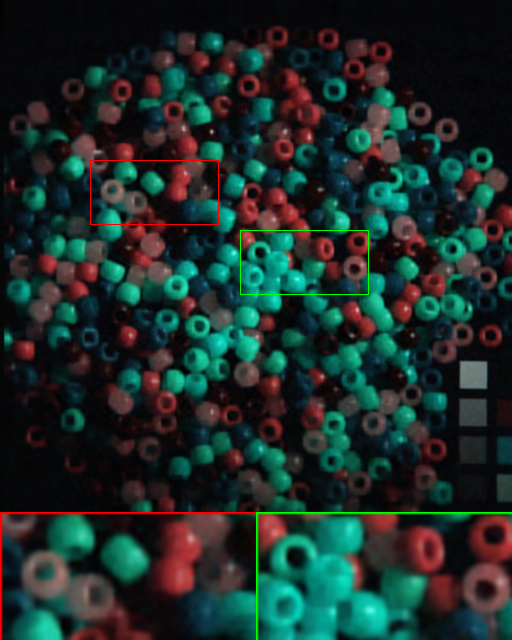}}
		\hspace{0.01cm}
		\subfloat[PSNR 27.18]
		{\includegraphics[width=0.115\linewidth]{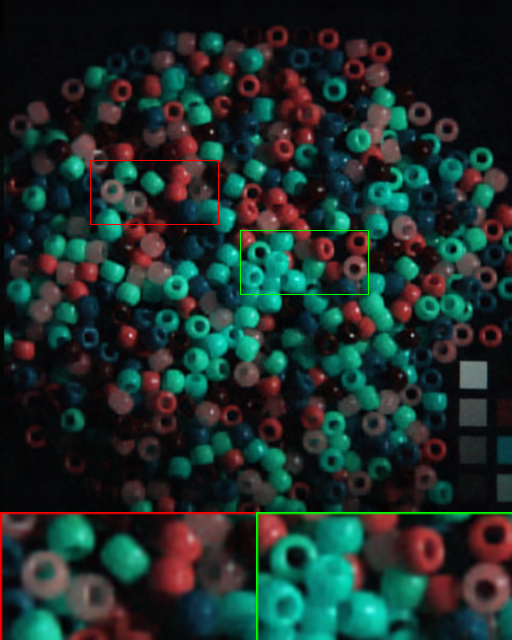}}
		\hspace{0.01cm}
		\subfloat[PSNR 27.20]
		{\includegraphics[width=0.115\linewidth]{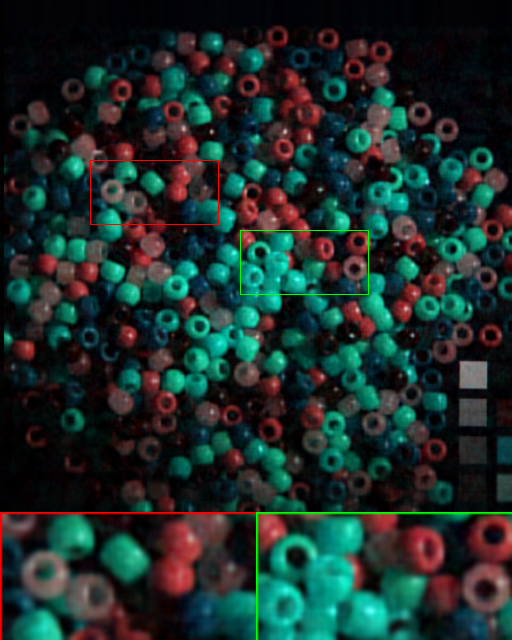}}
		\hspace{0.01cm}
		\subfloat[PSNR 26.08]
		{\includegraphics[width=0.115\linewidth]{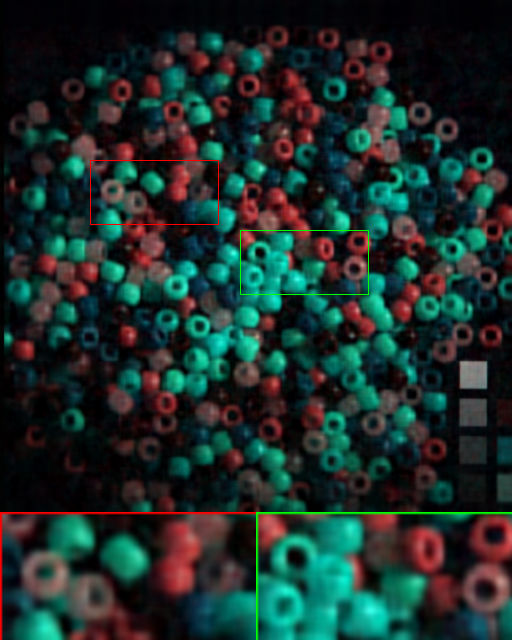}}
		\hspace{0.01cm}
		\subfloat[PSNR 27.93]
		{\includegraphics[width=0.115\linewidth]{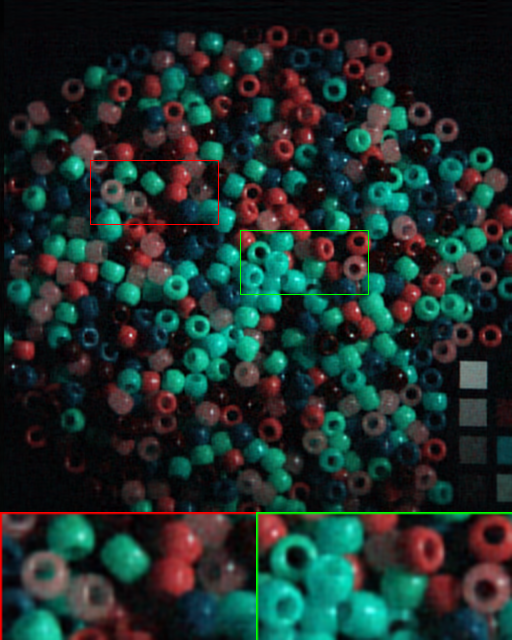}}
		\hspace{0.01cm}
		\subfloat[PSNR 28.52]
		{\includegraphics[width=0.115\linewidth]{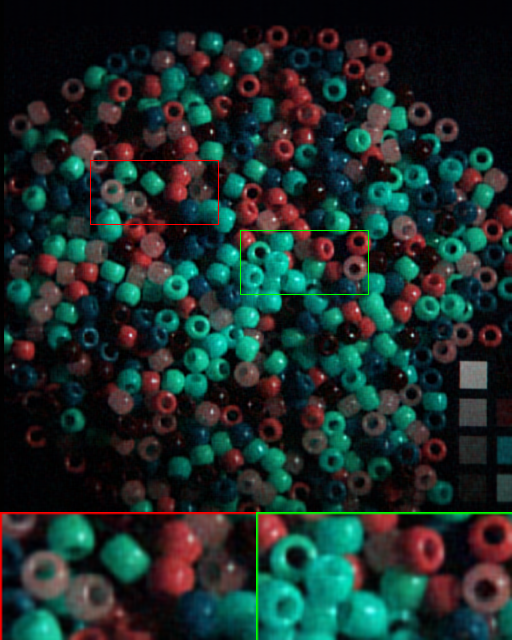}}
		\hspace{0.01cm}
		\subfloat[PSNR Inf]
		{\includegraphics[width=0.115\linewidth]{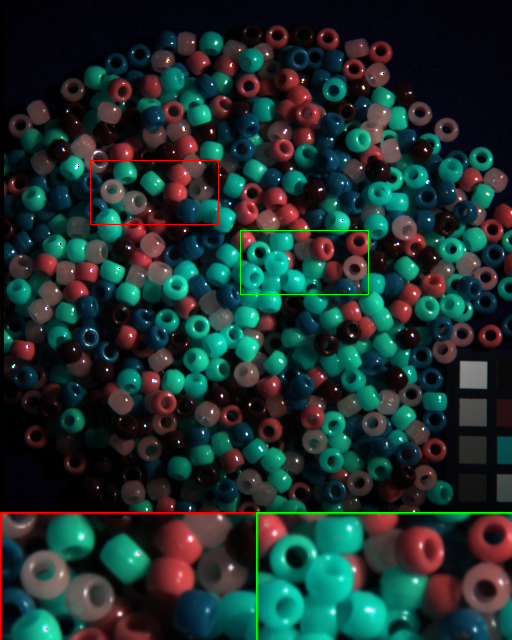}}
	\end{minipage}
	\begin{minipage}{1.\linewidth}
		\centering
		\subfloat[PSNR 15.90]
		{\includegraphics[width=0.115\linewidth]{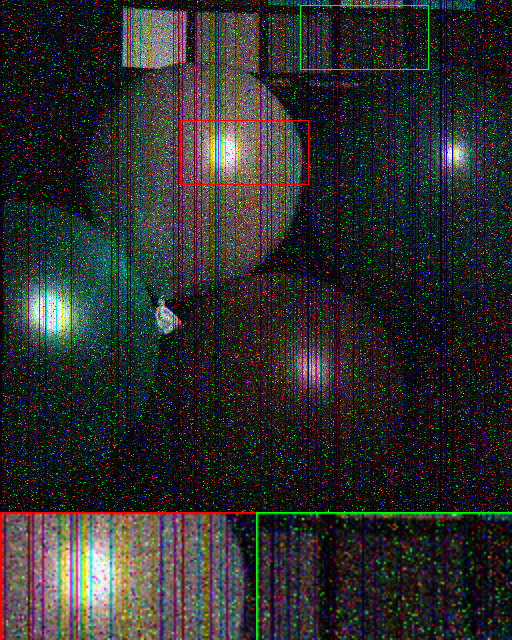}}
		\hspace{0.01cm}
		\subfloat[PSNR 29.89]
		{\includegraphics[width=0.115\linewidth]{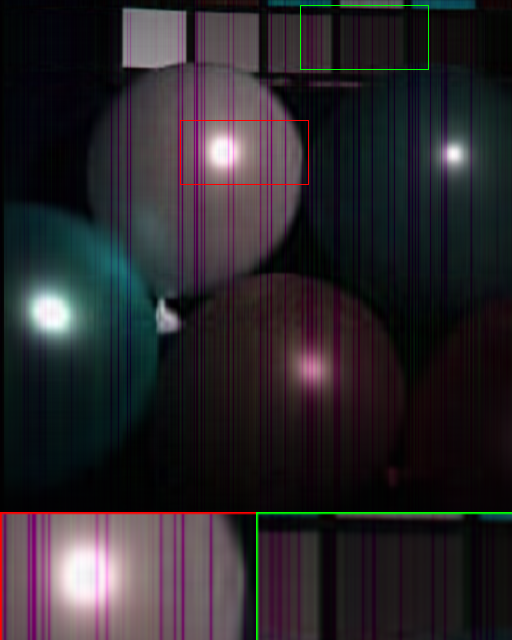}}
		\hspace{0.01cm}
		\subfloat[PSNR 32.06]
		{\includegraphics[width=0.115\linewidth]{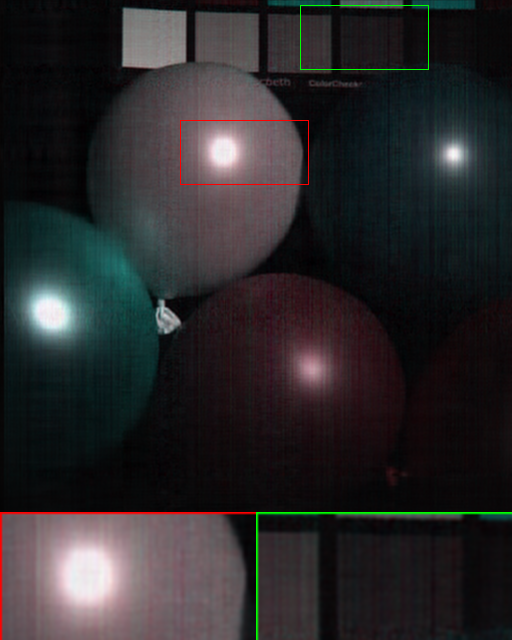}}
		\hspace{0.01cm}
		\subfloat[PSNR 32.41]
		{\includegraphics[width=0.115\linewidth]{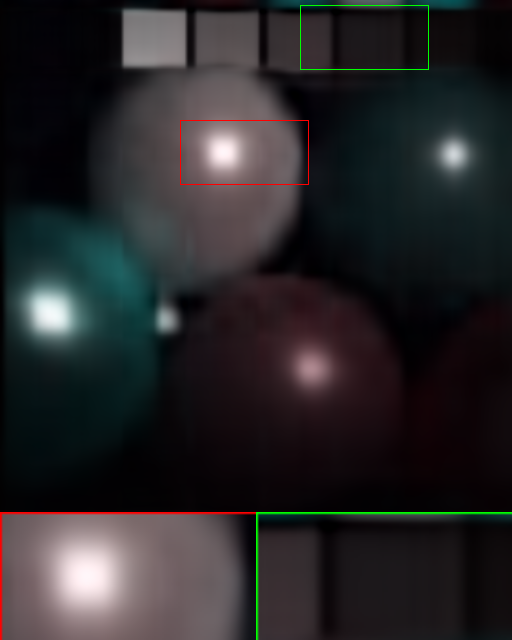}}
		\hspace{0.01cm}
		\subfloat[PSNR 30.75]
		{\includegraphics[width=0.115\linewidth]{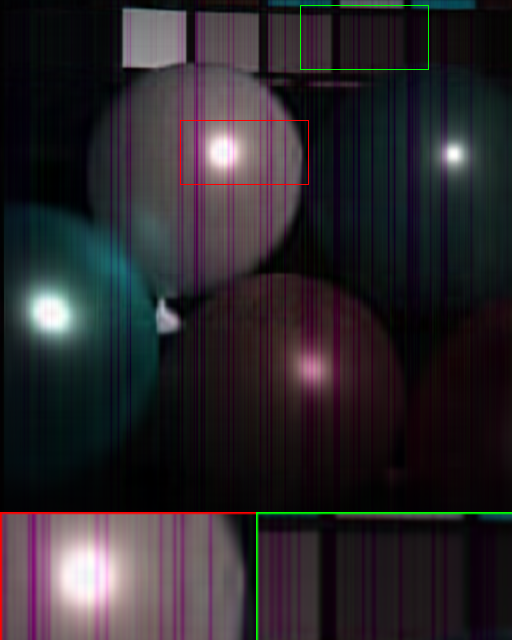}}
		\hspace{0.01cm}
		\subfloat[PSNR 33.80]
		{\includegraphics[width=0.115\linewidth]{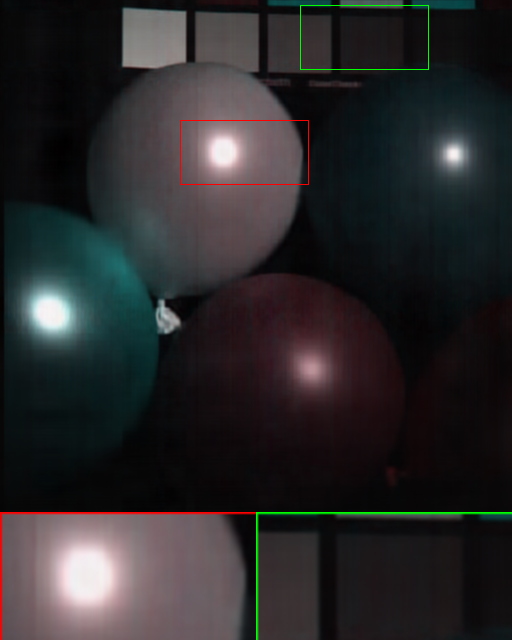}}
		\hspace{0.01cm}
		\subfloat[PSNR 34.43]
		{\includegraphics[width=0.115\linewidth]{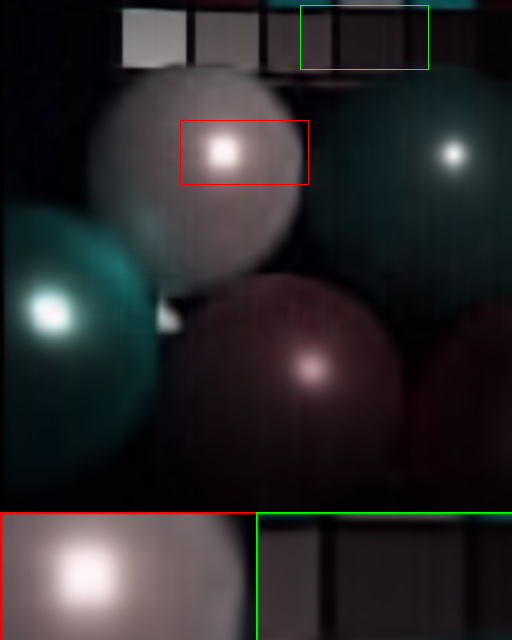}}
		\hspace{0.01cm}
		\subfloat[PSNR Inf]
		{\includegraphics[width=0.115\linewidth]{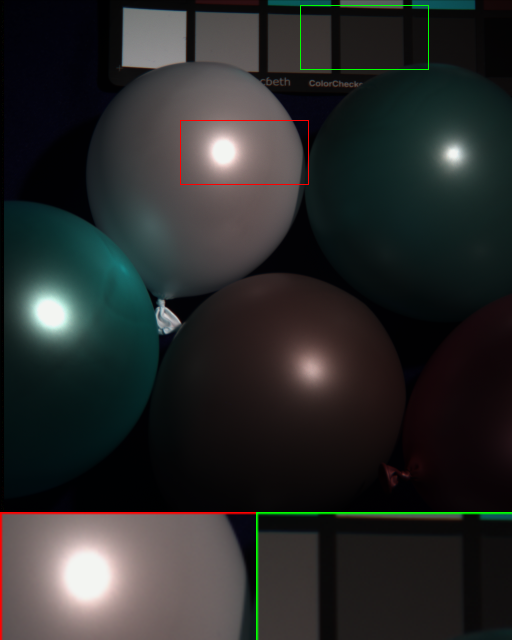}}
	\end{minipage}
	\begin{minipage}{1.\linewidth}
		\centering
		\subfloat[PSNR 16.49]
		{\includegraphics[width=0.115\linewidth]{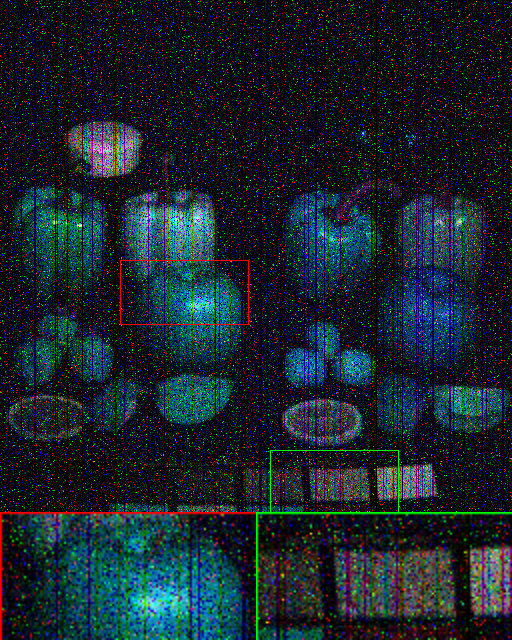}}
		\hspace{0.01cm}
		\subfloat[PSNR 28.55]
		{\includegraphics[width=0.115\linewidth]{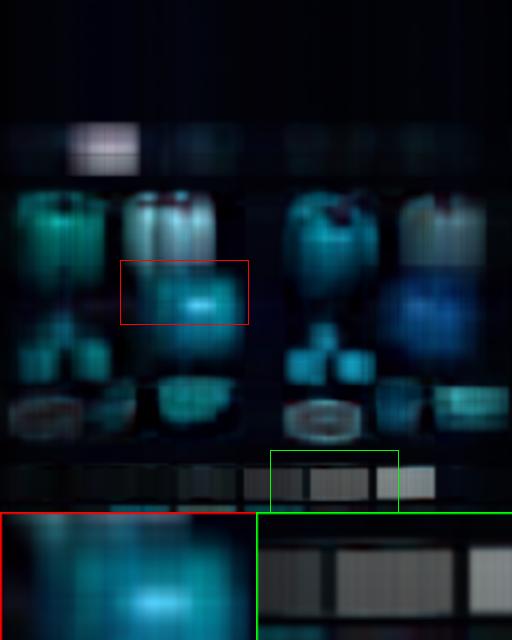}}
		\hspace{0.01cm}
		\subfloat[PSNR 29.93]
		{\includegraphics[width=0.115\linewidth]{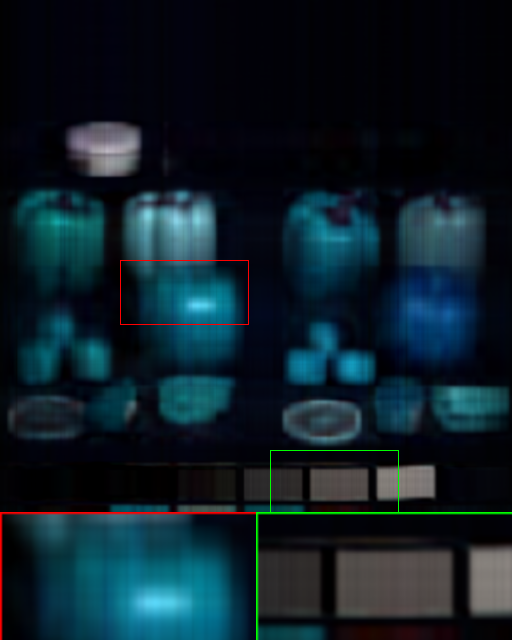}}
		\hspace{0.01cm}
		\subfloat[PSNR 30.65]
		{\includegraphics[width=0.115\linewidth]{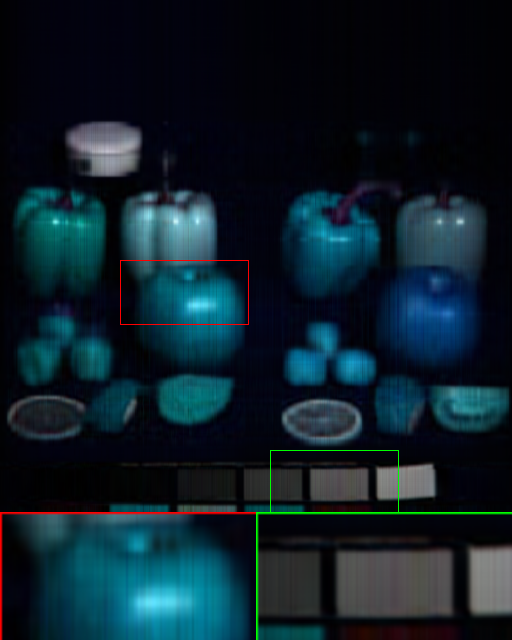}}
		\hspace{0.01cm}
		\subfloat[PSNR 29.51]
		{\includegraphics[width=0.115\linewidth]{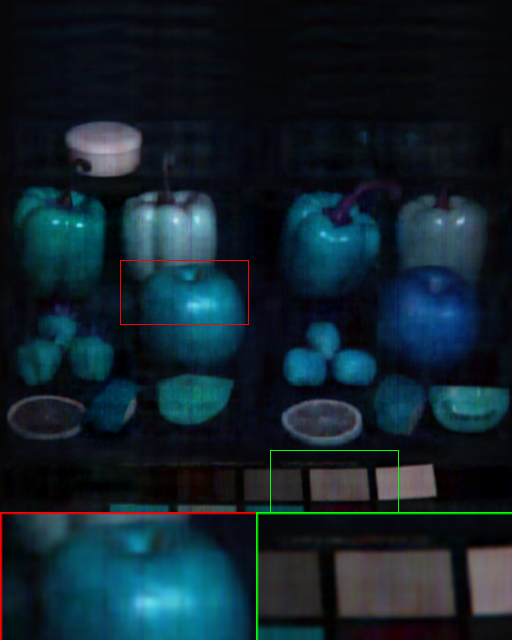}}
		\hspace{0.01cm}
		\subfloat[PSNR 31.42]
		{\includegraphics[width=0.115\linewidth]{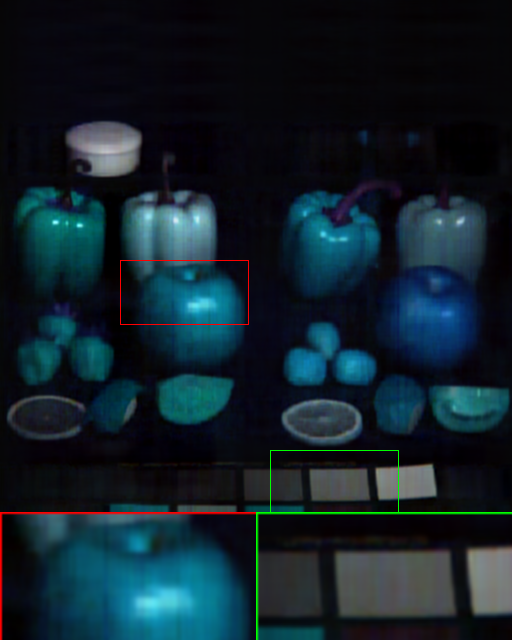}}
		\hspace{0.01cm}
		\subfloat[PSNR 31.85]
		{\includegraphics[width=0.115\linewidth]{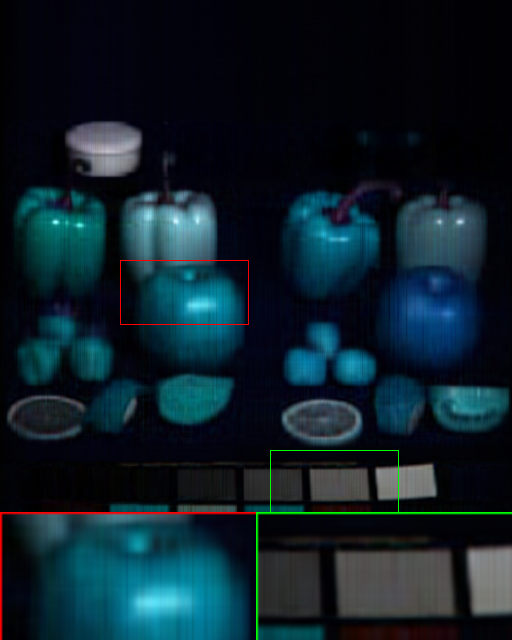}}
		\hspace{0.01cm}
		\subfloat[PSNR Inf]
		{\includegraphics[width=0.115\linewidth]{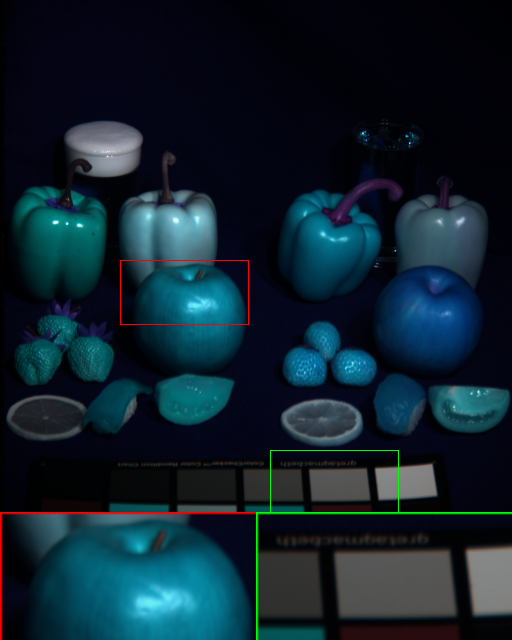}}
	\end{minipage}
	\begin{minipage}{1.\linewidth}
		\centering
		\subfloat[\begin{tabular}{c} PSNR 17.78\\ Observed\end{tabular}]
		{\includegraphics[width=0.115\linewidth]{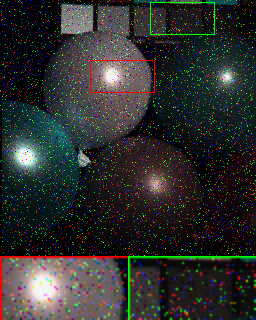}}
		\hspace{0.01cm}
		\subfloat[\begin{tabular}{c} PSNR 37.99\\ M\textsuperscript{2}DMT \end{tabular}]
		{\includegraphics[width=0.115\linewidth]{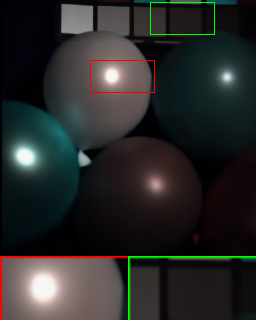}}
		\hspace{0.01cm}
		\subfloat[\begin{tabular}{c} PSNR 42.75\\ LRTC-ENR \end{tabular}]
		{\includegraphics[width=0.115\linewidth]{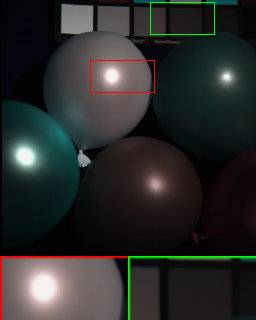}}
		\hspace{0.01cm}
		\subfloat[\begin{tabular}{c} PSNR 44.56\\ HLRTF \end{tabular}]
		{\includegraphics[width=0.115\linewidth]{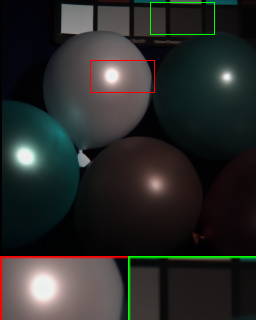}}
		\hspace{0.01cm}
		\subfloat[\begin{tabular}{c} PSNR 39.23\\ DeepTensor\end{tabular}]
		{\includegraphics[width=0.115\linewidth]{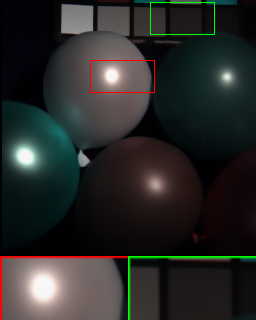}}
		\hspace{0.01cm}
		\subfloat[\begin{tabular}{c} PSNR 45.63\\ LRTFR\end{tabular}]
		{\includegraphics[width=0.115\linewidth]{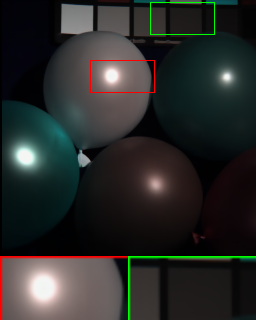}}
		\hspace{0.01cm}
		\subfloat[\begin{tabular}{c} PSNR 48.63\\ ScoreTR \end{tabular}]
		{\includegraphics[width=0.115\linewidth]{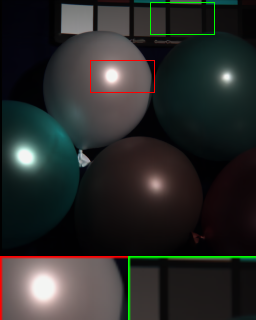}}
		\hspace{0.01cm}
		\subfloat[\begin{tabular}{c}PSNR Inf\\ Original \end{tabular}]
		{\includegraphics[width=0.115\linewidth]{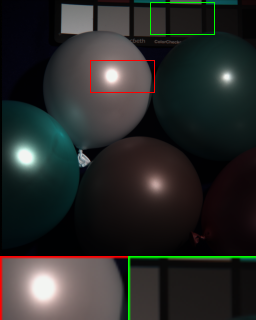}}
	\end{minipage}
	\caption{Results of multi-dimensional image denoising by different methods on multispectral images \textit{Balloons} (Case 1), \textit{Flowers} (Case 2), \textit{Beads} (Case 3), \textit{Balloons} (Case 4), \textit{Fruits} (Case 5) and \textit{Balloons} (Case 6).}
	\label{fig:Demo4Denoising}
\end{figure*}
The qualitative results of multispectral image denoising are shown in Figure~\ref{fig:Demo4Denoising}. 
Our method achieves the best performance across all six noise scenarios, demonstrating the strong blind source denoising capability of ScoreTR. This improvement is primarily attributed to the proposed smooth regularization. Unlike TV regularization, which requires a complete tensor as a prerequisite, our energy-based smoothing approach is both efficient and effective.

\end{document}